\documentclass{ecai} 

\usepackage{latexsym}
\usepackage{amssymb}
\usepackage{amsmath}
\usepackage{amsthm}
\usepackage{booktabs}
\usepackage{enumitem}
\usepackage{graphicx}
\usepackage{color}

\usepackage{bm}
\usepackage{arydshln}
\usepackage{multirow}
\usepackage{amssymb}
\usepackage{amsmath}
\usepackage{pifont}
\usepackage{xcolor}

\usepackage{subfigure}

\newcommand{\BibTeX}{B\kern-.05em{\sc i\kern-.025em b}\kern-.08em\TeX}

\newcommand{\cmark}{\ding{51}}%
\newcommand{\xmark}{\ding{55}}%

\begin{document}


\begin{frontmatter}


\paperid{123} 


\title{CRoC: Context Refactoring Contrast for Graph Anomaly Detection with Limited Supervision}





\author[1]{\fnms{Siyue}~\snm{Xie}\thanks{Corresponding Author.}}
\author[1]{\fnms{Da Sun Handason}~\snm{Tam}}
\author[1]{\fnms{Wing Cheong}~\snm{Lau}}
\address[1]{The Chinese University of Hong Kong}
\address{xiesiyue@link.cuhk.edu.hk, handasontam@gmail.com, wclau@ie.cuhk.edu.hk}


\begin{abstract}
Graph Neural Networks (GNNs) are widely used as the engine for various graph-related tasks, with their effectiveness in analyzing graph-structured data. 
However, training robust GNNs often demands abundant labeled data, which is a critical bottleneck in real-world applications.
This limitation severely impedes progress in Graph Anomaly Detection (GAD), where anomalies are inherently rare, costly to label, and may actively camouflage to evade detection. 
To address these problems, we propose \textbf{C}ontext \textbf{R}efact\textbf{o}ring \textbf{C}ontrast (CRoC), a simple yet effective framework that trains GNNs for GAD by jointly leveraging limited labeled and abundant unlabeled data.
Unlike previous works, CRoC exploits the class imbalance inherent in GAD to refactor the context of each node, which builds augmented graphs by recomposing the attributes of nodes while preserving their interaction patterns.
Furthermore, CRoC encodes heterogeneous relations separately and integrates them into the message-passing process, inducing the model to capture complex interaction semantics.
These operations preserve node semantics while encouraging robustness against adverse camouflage, enabling GNNs to uncover intricate anomalous cases.
In the training stage, CRoC is further integrated with the contrastive learning paradigm.
This allows GNNs to effectively harness unlabeled data during joint training, producing richer, more discriminative node embeddings. 
CRoC is evaluated on seven real-world GAD datasets with different sizes. 
Extensive experiments demonstrate that CRoC achieves up to $14\%$ AUC improvement over baseline GNNs and outperforms state-of-the-art GAD methods under limited-label settings.
\end{abstract}

\end{frontmatter}


\section{Introduction}

Graph Anomaly Detection (GAD) is a task aiming to identify nodes that deviate significantly from the majority in terms of roles or behaviors within a graph \cite{ma2021comprehensive}.
These instances, though rare, can have a profound impact on the entire system.
Therefore, GAD has drawn considerable attention across domains, with a wide range of applications, such as fraudster detection \cite{xiang2023semi}, malicious accounts mining \cite{xu2023self}, spam review filtering \cite{caregnn,graphconsis}, etc.

To detect graph anomalies, Graph Neural Networks (GNNs) \cite{gcn,graphsage} have emerged as a leading solution for GAD.
Generally, GNNs learn node embeddings by aggregating neighborhood information of the target node.
By iteratively stacking multiple layers, GNNs can encode useful context information critical for a downstream task.

However, there are still many obstacles when applying GNNs for GAD.
A primary concern is that GNNs are typically trained in a supervised manner, relying heavily on labeled data.
Training GNNs with very few labeled data may lead to severe overfitting and degraded performance.
However, in real-world GAD scenarios, informative labeled data are often scarce due to the fact that: (1) the target of interest, i.e., the anomalous nodes, are usually the minority in the graph, and (2) sometimes only experienced human experts are qualified to annotate, making it costly to collect adequate training labels.
Moreover, the class distribution in GAD is extremely imbalanced since normal nodes account for the majority.
Ignoring this nature may result in biased predictions.
Another concern is that typical GNNs work better in homophily graphs, where adjacent nodes are expected to be in the same class.
However, in GAD, it is common to find anomalous nodes connected to normal nodes, resulting in heterophilic connections.
This is because anomalous nodes, e.g., fraudsters in payment networks, may mimic normal nodes' features or interact with normal nodes to camouflage themselves.
Some previous works on GAD resort to pruning heterophilic edges \cite{ghrn,tam} or sampling homophilic edges \cite{caregnn,pcgnn} for GNN models.
However, such operations can be defective since GNNs are not inherently designed to react proactively against camouflage in the graph.

\begin{figure}[tb]
    \centering
    \includegraphics[width=\linewidth]{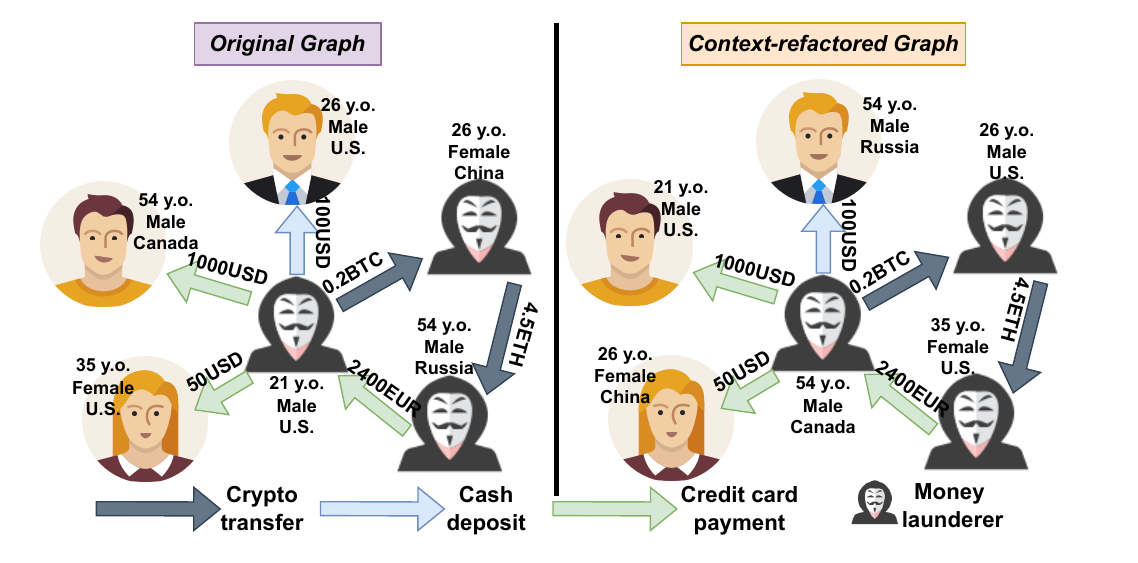}
    \caption{A toy example of context refactoring: randomly shuffle user profiles in a transaction network. The role of the centered anomalous node ought to be consistent if edge information is preserved.}
    \label{fig:croc_example}
\end{figure}

To address these challenges, we propose Context Refactoring Contrast (CRoC) to enhance GNNs for GAD when training labels are limited.
Considering the prevalence of camouflage in real-world GAD scenarios, we refactor the context of nodes by deliberately introducing (simulated) camouflage into the graph to train a more robust model.
(`Context' refers to any information about a node's neighborhood, including node/ edge attributes, graph topologies, etc.)
An illustrative example is shown in Fig. \ref{fig:croc_example}.
Concretely, we propose to refactor a graph by shuffling the node features but preserving the relations/ interactions of edges, which is expected to keep the semantics of a node consistent.
This is particularly effective for GAD scenarios since it utilizes the prior knowledge of class imbalance and (the existence of) camouflage in the graph.
Anomalous nodes with feature camouflage are observed to have node features similar to those of normal nodes \cite{caregnn}.
Since normal nodes account for the majority of the graph, shuffling node features hardly changes the semantics of normal and camouflaged anomalous nodes, as their context changes little.
On the other hand, anomalous patterns are highly correlated with edge information, which encodes special relations and interaction behaviors between nodes.
Preserving the edge information of the original graph for context refactoring essentially maintains most of the critical information required for GAD.
Through context refactoring, features of unlabeled nodes can also be reused in the context of the labeled nodes, which effectively correlates the labeled and unlabeled data.
In addition, we propose a relation-aware joint aggregation operator, enabling GNNs to distinguish diverse interaction patterns.
From the perspective of self-supervised learning, the context-refactored graph is an augmentation of the original graph in the view of semantics.
Therefore, it is natural to integrate context refactoring with contrastive learning.
A node-wise contrast is then applied between the original and the context-refactored graph.
With contrastive learning, all unlabeled nodes are actively incorporated into the training process, encouraging the model to learn from the abundant unlabeled data and yielding more comprehensive and robust node representations for GAD tasks.
We evaluate CRoC across seven real-world GAD datasets with only limited labels provided for training.
Extensive experiments show that CRoC can effectively enhance GNN models (up to $14\%$ improvement w.r.t AUC) and outperforms other state-of-the-art GAD schemes.
Our code\footnote{\url{https://github.com/XsLangley/CRoC_ECAI2025}} is publicly available. 


We summarize our contributions as follows:
\begin{itemize}
    \item We analyze some common challenges of GAD and propose context refactoring, which effectively correlates labeled and unlabeled data to train GNN against camouflage with limited supervision.
    \item We integrate context refactoring into the contrastive learning paradigm and propose a relation-aware joint aggregation operator, enabling a GNN to learn more comprehensive and discriminative representations for GAD tasks.
    \item We evaluate our scheme on seven real-world GAD datasets. Experimental results demonstrate CRoC's superiority when only limited labels are available for training.
\end{itemize}


\section{Related Works}

\subsection{Graph Neural Networks}


Graph Neural Networks (GNNs) are a family of models for dealing with graph-structured data.
Pioneer researchers design GNN kernels based on graph spectrum theory \cite{bruna2013spectral}, while follow-up works simplify the operator by a message-passing framework \cite{gcn,graphsage} and achieve better performances \cite{gin,gat}.
Similar to other deep-learning models, GNNs are typically trained in a supervised manner.
There are also works attempting to enhance GNNs by learning from unlabeled data.
One typical way is through self-training \cite{adaedge,m3s}, which picks confident predictions to extend the training set.
Another set of works pre-trains a GNN through contrastive learning \cite{bgrl,dgi}, where handcrafted augmentations are designed to form contrastive pairs for unsupervised training.
Others \cite{grand,cocos,violin} resort to training GNNs together with well-designed self-supervised objectives, aiming to align learned embeddings to a downstream task.

\subsection{Graph Anomaly Detection with GNNs}

With great effectiveness in processing graph data, GNNs are elaborately adapted to handle GAD tasks as per different challenges.
Some works utilize GAD's class imbalance (i.e., normal nodes are the majority of the graph) for detecting anomalies.
They train GNNs to reconstruct a graph, where normal nodes are assumed to be better reconstructed.
Instances with relatively poor reconstruction results are flagged as anomalous \cite{gad-nr,ada-gad}.
Some methods rely on the proximity hypothesis to train their detector, where neighboring nodes' representations are assumed to be similar to each other.
A node is supposed to be anomalous if its representation is dissimilar to its neighbors' \cite{tam,anemone}.
Researchers also notice that camouflage will result in non-homophilic connections, which are undesired for typical GNNs.
To deal with it, several approaches \cite{caregnn,dagnn,pcgnn} propose selective node sampling during aggregation to reduce heterophilic interference.
BWGNN \cite{bwgnn} and SplitGNN \cite{splitgnn} handle GAD by applying well-designed graph kernels with tunable frequencies.
GHRN \cite{ghrn} prunes heterophilic edges to build a new graph for training, while GDN \cite{gdn} and GTAN \cite{gtan} mine heterophily-robust features for the model.
Another group of methods \cite{dci,consisgad,bsl} attempts to enhance GNNs by introducing graph augmentations and self-supervised learning techniques.
However, many aforementioned works assume that they have abundant labeled anomalous data to train their GNN backbones or specialized modules, such as the node-wise similarity predictor in CARE-GNN or the edge heterophily indicator in GHRN.
The effectiveness of these modules can be challenged when labels for training are very limited, which is common in real-world GAD scenarios.
More discussions on related works can be found in \ref{app:rel_work}.

\section{Preliminary Analysis and Motivations}

\subsection{Problem Formulations}

We formulate GAD as a semi-supervised node classification task: given a multi-relation graph $\mathcal{G} = (\mathcal{V}, \bm{X}, \{\mathcal{E}_{r}\}|_{r=1}^{R})$, where $\mathcal{V} = \{ v_i\}|_{i=1}^{N}$ is a $N$-sized node set and $\bm{X}=[\bm{x}_{1},...,\bm{x}_{N}]^{\intercal}\in \mathbb{R}^{N\times d_{0}}$ is the corresponding node feature matrix.
The edge set $\{\mathcal{E}_{r}\}|_{r=1}^{R}$ encompasses $R$ relations.
Each edge $e_{i,j}^{r}=(v_i, v_j)\in \mathcal{E}_{r}$ is associated with a relation $r$.
$\mathcal{V}$ consists of two subsets $\mathcal{V} = \{ \mathcal{V}^{L}, \mathcal{V}^{U} \}$, which denotes the labeled ($\mathcal{V}^{L}$) and unlabeled ($\mathcal{V}^{U}$) node set respectively.
Let $\mathbf{Y}^{L} = \{ \bm{y}_{v}| v \in \mathcal{V}^{L} \}$ be the labels for $\mathcal{V}^{L}$, where $\bm{y}_{v}$ is a one-hot vector to indicate the class of node $v$.
Note that in GAD with limited supervision, $|\mathcal{V}^{L}| \ll N$.
Our goal is to train a GNN-based model given $\mathcal{G}$ and $\mathbf{Y}^{L}$ to predict the class (normal/ anomalous) of all unlabeled nodes.

\subsection{Challenges of Camouflage in Graphs}

\label{sect:preli:camouflage}

Camouflage poses a significant challenge in GAD tasks, which has drawn great attention in some previous works \cite{caregnn,fraudre}.
Typically, there are two kinds of camouflage in GAD.
One is node feature camouflage, where anomalous nodes deliberately mimic their node features like normal nodes.
For example, fraudsters may declare false personal information in social networks, rendering their profile (node features) less reliable for GAD.
Another type is behavior camouflage, where anomalous nodes engage in both malicious and benign interactions simultaneously.
Such malicious interactions can be informative for GAD.
However, these critical clues may be overwhelmed by accompanying normal interactions if all interactions are equally treated.
Therefore, to avoid being misled by camouflage, we should train a GNN to be robust against feature camouflage and endow it with the ability to distinguish different kinds of interactions in the graph.

\subsection{Limited Supervision for GNNs in GAD Tasks}

\begin{table}[tb]
    \centering
    \caption{Statistics of two typical GAD datasets.}
    \begin{tabular}{c|cccc}
        \hline
         Dataset    &  \#Node   &  \#Edge       &   \#Anomaly   &   Anomaly(\%)\\
         \hline
         T-Soc      &  5.7M   &  73M   &   174,280     &   3.01\%   \\
         DGraph     &  3.7M   &  4.3M    &   15,509      &   0.42\%   \\
         \hline
    \end{tabular}
    \label{tab:preli:ds_stat}
\end{table}

GNNs' performance is highly correlated with the amount of labeled data provided for training.
However, in real-world scenarios, supervised information (labels) is very limited due to the great difficulty and high expense of annotation.
In GAD, this challenge is exacerbated by the naturally imbalanced class distribution, where our target of interest, i.e., anomalous nodes, are always the minority.
We show the statistics of two real-world GAD datasets in Table \ref{tab:preli:ds_stat} as an example.
Suppose $1\%$ of nodes could be annotated for training, the desired supervised information (labeled anomalies) remains scarce even for large-scale datasets.
For example, in DGraph, only around 155 anomalous nodes can be labeled for training in this case.
In other words, GNNs can only learn very limited knowledge about anomalies from labeled data, leading to inferior representations for GAD.

\begin{table}[tb]
    \centering
    \caption{Averaged node coverage of a typical GNN on \underline{T-Soc} with variant depth and label rate settings.}
    \begin{tabular}{c|ccc}
    \hline
       label rate  & 0.01\% & 0.05\% & 0.1\% \\
       \hline
        1-layer & 0.25\% & 1.25\% & 2.48\% \\
        2-layer & 8.73\% & 22.12\% & 31.47\% \\
        3-layer & 33.82\% & 56.53\% & 67.77\% \\
    \hline
    \end{tabular}
    \label{tab:preli:nodecov}
\end{table}

On the other hand, many unlabeled nodes remain underutilized when training GNNs following standard semi-supervised learning paradigms \cite{gcn,gat}.
To quantify this, \textbf{we define the node coverage of training as: $\frac{N_{L}^{k}}{N}$, where $N_{L}^{k}$ is the number of nodes within $k$-hop/layer neighbourhood of all labeled training nodes.}
Table \ref{tab:preli:nodecov} shows the statistics of typical message-passing GNNs (e.g., GCN and GAT).
We can observe that less than $70\%$ of nodes participate in training, even if we stack a 3-layer GNN model with a $0.1\%$ label rate setting (which is equivalent to providing up to 5781 labeled nodes for training). 
Coverage drops significantly if given lower annotation budgets.
In other words, many unlabeled nodes are essentially underexplored.
However, there can be useful patterns that occur only within unlabeled nodes.
The training data can hardly match the distribution of all data with such limited supervision, which may make it fail to generalize to the testing environment.
Therefore, besides training GNNs with the labeled data, we should manage to utilize the abundant unlabeled data to learn more comprehensive node embeddings for better generalization ability.


\section{Proposed Method}

\begin{figure*}[t]
    \centering
    \includegraphics[width=2\columnwidth]{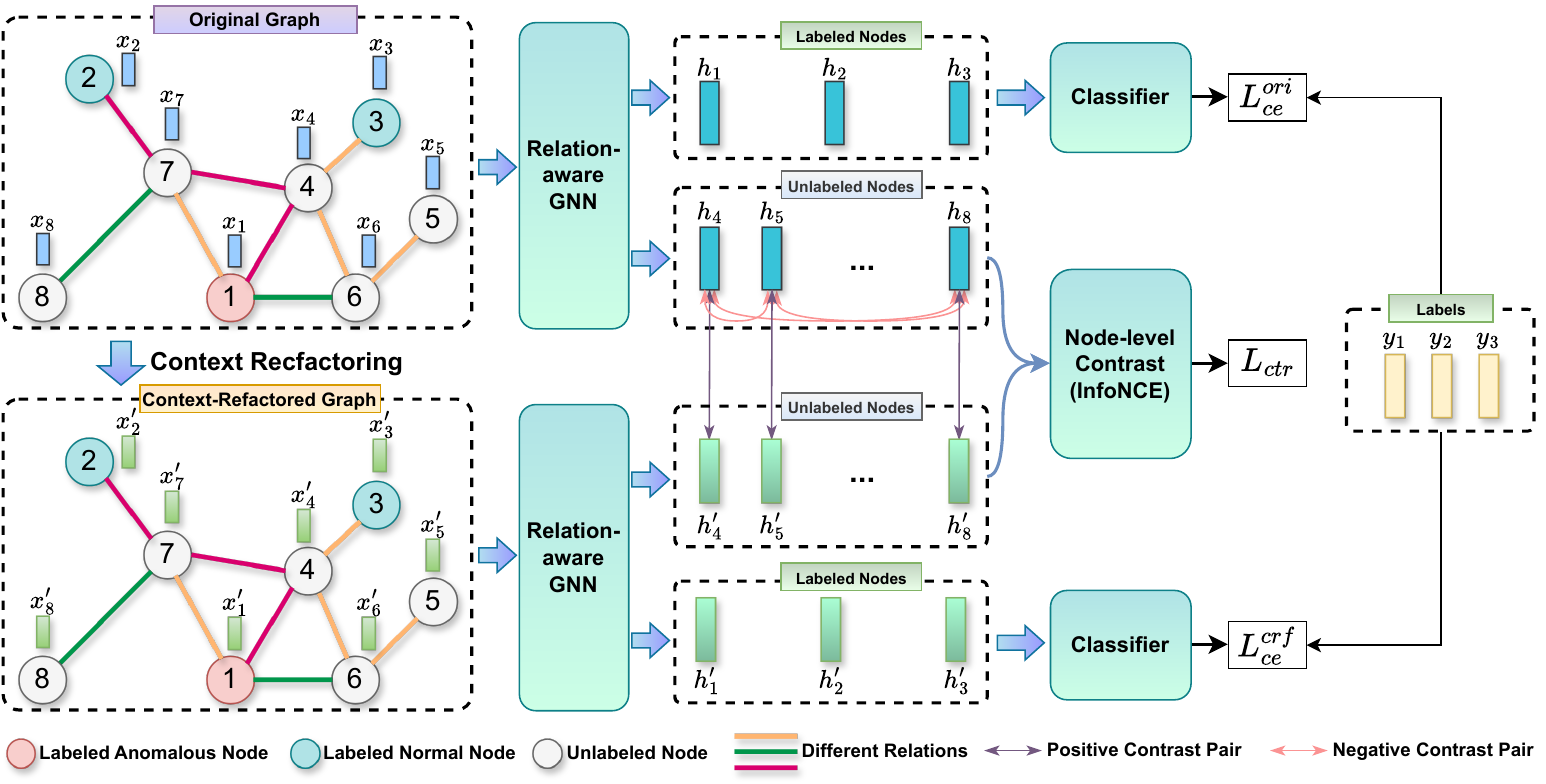}
    \caption{The framework of \textbf{C}ontext \textbf{R}efact\textbf{o}ring \textbf{C}ontrast (CRoC). 
    Given a multi-relation graph, we refactor the original graph utilizing the class imbalance prior of GAD. 
    Embeddings of the few labeled nodes from two different views (graphs) are supervised by ground-truth annotations, while unlabeled nodes are included for joint training via a self-supervised objective.
    }
    \label{fig:method:framework}
\end{figure*}

To deal with the challenges discussed in the previous section, we propose \textbf{C}ontext \textbf{R}efact\textbf{o}ring \textbf{C}ontrast (CRoC) for GAD tasks under limited supervision.
The framework of CRoC is shown in Fig. \ref{fig:method:framework}.

\subsection{Context Refactoring for GAD}


As discussed in the previous section, node feature camouflage may mislead the model when making decisions based on node features.
Previous works have attempted to detect such camouflage by measuring the feature similarity or disparity between node \cite{caregnn,dagnn}.
However, this is unreliable as features of anomalous nodes can be the same as normal nodes, e.g., a money launderer may deliberately declare his/her user information (node features) the same as others.
This easily collapses feature-based detectors, where we show an illustrative example in the visualization results (in Fig. \ref{fig:exp:case_viz}).
In this work, instead of endeavoring to detect camouflages, we go in the opposite direction: we proactively introduce node feature camouflage to refactor the graph.
We train a GNN on the refactored graph, forcing it to learn to be robust against node feature camouflage.

To this end, an intuitive way is to replace the features of a labeled anomalous node with those of normal nodes.
In CRoC, we adopt a more comprehensive strategy by refactoring the context of each node.
Formally, we define the context refactoring as follows:
\begin{equation}
    \bm{\tilde{X}} =  \Omega(\bm{X})
    \label{equ:method:feat_shuf}
\end{equation}
where $\Omega(\cdot)$ is a random shuffle function that permutes the original feature matrix $\bm{X}$ by row (i.e., we exchange the row vectors in the matrix), and $\bm{\tilde{X}}\in\mathbb{R}^{N\times d_{0}}$ is the feature matrix of the context-refactored graph.
We argue that Equ (\ref{equ:method:feat_shuf}) generally keeps the semantics of most nodes, which is especially effective for GAD by explicitly leveraging the class imbalance premise
.
On the one hand, normal nodes always account for the majority in a graph.
Therefore, a normal node will be assigned node features of the other normal node with a high probability, which has minimal impact on its overall contextual semantics.
On the other hand, undesired shuffle results (i.e., a normal node is assigned an anomalous node's features) are tolerable for a GNN.
This is because GNNs rely not only on the information of the target node but also on its neighborhood context for decision-making.
In our experiments, features will also be reshuffled in each training epoch, preventing the case where a normal node is always assigned undesired features.
For anomalous nodes, they are assigned normal nodes' features with a high probability, simulating scenarios where node feature camouflage occurs.
Training a GNN on a graph with explicit camouflage will force the model to learn to be more robust against camouflage. 
Therefore, instead of designing specific modules to alleviate the impact of node feature camouflage and class imbalance, we proactively harness these priors to boost the model.

Furthermore, considering cases where native node features are critical for detection, we propose mixing the original and shuffled node feature matrices to enhance the generalization ability:
\begin{equation}
    \bm{X'} = \alpha\bm{X} + (1 - \alpha)\bm{\tilde{X}}
    \label{equ:method:feat_mix}
\end{equation}
where $\alpha\in[0, 1)$ is a hyper-parameter to tune the importance of native node features for context refactoring.
The finalized context-refactored graph is: $\mathcal{G}' = (\mathcal{V}, \bm{X'}, \{\mathcal{E}_{r}\}|_{r=1}^{R})$.
This Mixup \cite{mixup} style augmentation enables CRoC to preserve node-specific identity information while introducing controllable variations to the graph.
More importantly, unlike the original Mixup scheme, our method, as shown in Equ (\ref{equ:method:feat_mix}), actively incorporates unlabeled nodes into training, enabling the GNN backbone to learn more diverse patterns in GAD.
More discussions refer to \ref{app:rel_work:comp_rel_work} and \ref{app:discussion:lb_ulb}.

\subsection{Relation-aware Joint Aggregation (RJA)}




In context refactoring, we shuffle node features but preserve the interaction connections (i.e., relations and edges), as interactions reflect critical anomalous patterns.
However, as discussed in Section \ref{sect:preli:camouflage}, anomalous instances may conceal malicious interactions within a multitude of benign interactions for behavior camouflage.
To handle it, some works propose to aggregate information from each relation individually within each layer \cite{caregnn,dagnn} or decouple the multi-relation graph into multiple single-relation graphs \cite{bwgnn}.
However, when a malicious activity involves multiple types of interactions/ relations, these schemes may disrupt the chain of interactions or lose track of some critical anomalous patterns.
An example is shown in Fig. \ref{fig:croc_example}, where the money laundry chain involves two payment schemes.

To address this problem, we propose a Relation-aware Joint Aggregation (RJA) operator to learn across different relations/ interactions in multi-relation graphs.
Specifically, we learn a relation embedding for each relation type in each GNN layer.
Denote $\bm{h}^{l}_{r_{(u,v)}}$ as the learnable relation embedding corresponding to edge $(u,v)$ in the $l$-th GNN layer, where $r_{(u,v)} \in \{ 1,2,...,R\}$ indexes the relation type of edge $(u,v)$.
We reformulate the aggregation operator of a message-passing GNN as follows:
\begin{align}
    \label{equ:method:edge_emb}
    \tilde{\bm{h}}^{l}_{u} &= \sigma\left( \bm{W}^{l} (\bm{h}^{l-1}_{u}\Vert \bm{h}^{l}_{r_{(u,v)}}) + \bm{b}^{l} \right) \\
    \label{equ:method:aggregation}
    \bm{h}^{l}_{v} &= \Phi\left( \bm{h}^{l-1}_{v}, Aggr(\{ \tilde{\bm{h}}^{l}_{u} \}) ; u\in\mathcal{N}_{v}  \right)
\end{align}
where $\bm{h}^{l}_{v}$ is the embedding of node $v$ in the $l$-th layer, $\Vert$ is the concatenation operator, $\bm{W}^{l}$ and $\bm{b}^{l}$ are learnable weight and bias in the $l$-th layer.
$\sigma(\cdot)$, $Aggr(\cdot)$ and $\Phi(\cdot)$ are activation, aggregation and transformation functions, which vary across different GNN backbones.
$\mathcal{N}_{v}$ is the neighbor set of the target node $v$, with neighbor $u$ connected via relation type $r_{(u,v)}$.
Equ (\ref{equ:method:edge_emb}) explicitly associates a relation with the neighbor interacting with the target node, enabling a GNN to investigate cross-relational information via Equ (\ref{equ:method:aggregation}).
RJA explicitly distinguishes different types of relations and encourages a GNN to learn mutual correlations among them.
Experimental results (Section \ref{sect:exp:abs:module} and \ref{sect:exp:qua_ana}) also demonstrate its effectiveness.

\subsection{Learn from the Context-Refactored Graph}

Given that the semantics of the context-refactored graph should align with the original graph, we can train a GNN in a supervised manner on both graphs using the available labels.
Denote $g(\mathcal{G}^{*}; \theta_{G})$ as a GNN parameterized by $\theta_{G}$ with an input graph $\mathcal{G}^{*}$ and $f(\bm{H}^{*}; \theta_{C})$ as the classifier parameterized by $\theta_{C}$ with input $\bm{H}^{*}$.
In our experiments, $g(\mathcal{G}^{*}; \theta_{G})$ is implemented by GIN \cite{gin} or GraphSAGE \cite{graphsage}, while a 2-layer MLP is used for $f(\bm{H}^{*}; \theta_{C})$.
Node embeddings from two views can be generated by feeding these two graphs (the original and the context-refactored graph) into the same GNN:
\begin{equation}
    \label{equ:method:emb}
    \bm{H} = g(\mathcal{G}~; \theta_{G}), ~~ \bm{H}' = g(\mathcal{G}'; \theta_{G})
\end{equation}
where $\bm{H}=[\bm{h}_{1}, \bm{h}_{2}, ..., \bm{h}_{N}]^{\intercal}$ and $\bm{H}'=[\bm{h}'_{1}, \bm{h}'_{2}, ..., \bm{h}'_{N}]^{\intercal}$ are $\mathbb{R}^{N\times d}$ embedding matrices w.r.t the original and the refactored graph, with $d$ the embedding dimension.
The classifier's predictions are formulated as:
\begin{equation}
    \label{equ:method:pred}
    \hat{\bm{Y}} = f(\bm{H}; \theta_{C}), ~~ \hat{\bm{Y}}' = f(\bm{H}'; \theta_{C}) \\     
\end{equation}
where $\hat{\bm{Y}}=[\hat{\bm{y}}_{1}, \hat{\bm{y}}_{2}, ..., \hat{\bm{y}}_{N}]^{\intercal}$ and $\hat{\bm{Y}}'=[\hat{\bm{y}}'_{1}, \hat{\bm{y}}'_{2}, ..., \hat{\bm{y}}'_{N}]^{\intercal}$ represent the normalized prediction confidence matrices.
We can formulate the objectives over the set of labeled nodes by the cross-entropy loss:
\begin{align}
    \label{equ:method:ori_loss}
    L^{ori}_{ce} &= -\frac{1}{|\mathcal{V}^{L}|} \sum_{v\in\mathcal{V}^{L}}\bm{y}_{v}^{\intercal}\log{(\hat{\bm{y}}_{v})} \\
    \label{equ:method:crf_loss}
    L^{crf}_{ce} &= -\frac{1}{|\mathcal{V}^{L}|} \sum_{v\in\mathcal{V}^{L}}\bm{y}_{v}^{\intercal}\log{(\hat{\bm{y}}'_{v})}
\end{align}
By jointly optimizing Equ (\ref{equ:method:ori_loss}) and (\ref{equ:method:crf_loss}), we can train a GNN in a supervised manner.

It is worth noting that $L^{crf}_{ce}$ enables GNNs to access the abundant unlabeled data and correlate it with the limited labeled data during training.
This is because features of unlabeled nodes can be shuffled to the neighborhood of a labeled node, thereby contributing to $L^{crf}_{ce}$ by mixing with the original features by Equ (\ref{equ:method:feat_mix}).
This operation enables GNNs to efficiently reuse the context information of labeled nodes (by mixing up different node features). 
This improves the input diversity and implicitly reduces the risk of overfitting limited labeled data.
Moreover, unlike methods that learn from unlabeled data via pseudo-labels \cite{consisgad,m3s}, $L^{crf}_{ce}$ is fully supervised by ground-truth labels.
This avoids the pitfalls of learning from incorrect knowledge (raised by noisy pseudo-labels) and thereby generates more discriminative embeddings.
More analysis refers to \ref{app:discussion:lb_ulb}.

\subsection{Context Refactoring Contrast}


From the perspective of data augmentation, context refactoring is a graph augmentation scheme specific to GAD tasks.
This makes it a natural fit for integration with contrastive learning paradigms.

Specifically, we apply a node-wise contrast scheme for training.
As shown in Fig. \ref{fig:method:framework}, we collect unlabeled nodes' embeddings from the two different views (the original and the context-refactored graphs).
Since context refactoring preserves the context semantics, we pair an unlabeled node's embedding from the original graph with its counterpart from the context-refactored graph to form a positive sample.
For negative samples, we use an intra-view approach by randomly selecting embeddings of different nodes from the original graph. 
We finally contrast the positive against negative samples following the InfoNCE loss \cite{moco}:
\begin{equation}
    \label{equ:method:ctr_loss}
    L_{ctr} = - \frac{1}{|\mathcal{V}^{U}|}\sum_{v, u_{i}\in\mathcal{V}^{U}}\log \frac{exp(\bm{h}_{v}^{\intercal}\bm{h}_{v}' / \tau)}{\sum_{u_{i}\neq v}^{ i=1\sim k}exp(\bm{h}_{v}^{\intercal}\bm{h}_{u_{i}} / \tau)}
\end{equation}
where $\tau$ is a tunable temperature and $k$ is the number of negative samples for each positive sample.

Note that Equ (\ref{equ:method:ctr_loss}) incorporates all unlabeled nodes in the objective function.
This enables a GNN to learn from the neighborhood context of the massive unlabeled nodes, inducing the model to capture more diverse patterns for generating more powerful node embeddings.
Additional analyses refer to \ref{app:discussion:diverse_pattern}.

\subsection{Train GNNs with CRoC}

We enhance a GNN-based GAD system via CRoC by optimizing the following loss function:
\begin{equation}
    \label{equ:method:overall_loss}
    L_{all} = L^{ori}_{ce} + \gamma L^{crf}_{ce} + \eta L_{ctr}
\end{equation}
where $\gamma$ and $\eta$ are two hyper-parameters to balance the objective.
Note that in Equ (\ref{equ:method:overall_loss}), we unify the self-supervised objective ($L_{ctr}$) and the supervised parts ($L^{ori}_{ce}$ and $L^{crf}_{ce} $) for joint training in an end-to-end manner.
Therefore, more diverse patterns can be learned via $L_{ctr}$ to supplement the limited supervised information for decision making.
Simultaneously, $L_{ctr}$ is regulated by the supervised signal provided by $L^{ori}_{ce}$ and $L^{crf}_{ce}$, ensuring the model to learn extra knowledge that will not deviate from the target of the task.
Therefore, CRoC bridges the few labeled data and abundant unlabeled data, enabling GNNs to learn more powerful representations for GAD tasks under the guidance of scarce supervised signals.



\section{Experimental Evaluations}

\subsection{Experimental Setup}

\subsubsection{Datasets}

We evaluate CRoC across seven real-world GAD datasets, encompassing diverse domains such as commodity review fraud (Amazon and Yelp), financial fraud accounts (T-Fin and DGraph) and anomalies in social networks (T-Soc, Weibo and Reddit).
Detailed descriptions and statistics for all datasets are provided in \ref{app:exp_setting:dataset}.

\subsubsection{Baselines}

We compare CRoC with three groups of methods for GAD tasks:
\textbf{(1) Common feature-based classifiers:} MLP and XGBoost (XGB) \cite{xgboost}. 
\textbf{(2) GNN baselines:}  GCN \cite{gcn}, GAT \cite{gat}, GIN \cite{gin}, SAGE \cite{graphsage} and RGCN \cite{rgcn}. 
\textbf{(3) Specialized GAD models:} CARE-GNN \cite{caregnn}, DCI \cite{dci}, PCGNN \cite{pcgnn}, BWGNN \cite{bwgnn}, XGBGraph \cite{gadbench}, GHRN \cite{ghrn} and ConsisGAD \cite{consisgad}.
Methods \cite{fraudre,gdn,dagnn} that do not report better results than the listed models are omitted for comparison.



\subsubsection{Implementations}

We follow BWGNN's settings \cite{bwgnn} to split datasets: $1\%$ ($0.01\%$ for T-Soc) of nodes are randomly sampled to form the training set (labeled nodes), while the remaining are split into validation and test sets in a proportion of $1:2$.
We implement CRoC by a 2-layer GIN (or SAGE) with 64 hidden dimensions.
We set $\tau=2$, search $\alpha$, $\gamma$ and $\eta$ in $[0, 1]$ and train GNNs with learning rate $0.003$.
In experiments, we build a new context-refactored graph for every training epoch.
The model is finally evaluated on the original graph.
We run each model 10 rounds with different random seeds and record the test result w.r.t. the best validation results for each round.
We use the Area under the ROC Curve (AUC) and the Average Precision score (AP) to measure performance and the reported value is averaged over ten runs.
More details refer to \ref{app:exp_setting:imp_detail}.

\subsection{Results on GAD with Limited Supervision}

\label{sect:exp:results}

\begin{table*}
    \centering
    \caption{Experimental Results on five GAD Datasets. The label rate of each dataset is marked in parentheses.}
    \resizebox{2\columnwidth}{!}{
    \begin{tabular}{c|cc|cc|cc|cc|cc}
        \hline
         Dataset & \multicolumn{2}{c|}{T-Soc (0.01\%)} & \multicolumn{2}{c|}{DGraph (1\%)} & \multicolumn{2}{c|}{Yelp (1\%)} & \multicolumn{2}{c|}{T-Fin (1\%)} & \multicolumn{2}{c}{Amazon (1\%)} \\
         \hline
         Metric & AUC & AP & AUC &  AP & AUC & AP & AUC &  AP & AUC & AP \\
         \hline
         MLP & $69.49_{\pm3.06}$ & $6.96_{\pm1.29}$ & $71.20_{\pm0.24}$ & $2.55_{\pm0.06}$ & $72.43_{\pm0.85}$ & $31.61_{\pm1.71}$ & $90.02_{\pm1.12}$ & $58.47_{\pm2.94}$ & $89.20_{\pm1.39}$ & $53.85_{\pm5.78}$ \\
         XGB & $66.67_{\pm2.04}$ & $6.35_{\pm0.98}$ & $68.79_{\pm0.40}$ & $2.39_{\pm0.05}$ & $75.96_{\pm1.13}$ & $38.20_{\pm2.29}$ & $92.39_{\pm0.50}$ & $73.96_{\pm3.48}$ & ${91.29}_{\pm4.31}$ & ${72.83}_{\pm10.75}$ \\
         \hdashline
         GCN & $81.85_{\pm2.07}$ & $18.58_{\pm6.32}$ & $69.25_{\pm1.09}$ & $2.55_{\pm1.09}$ & $55.81_{\pm1.15}$ & $18.23_{\pm1.21}$ & $91.55_{\pm0.66}$ & $73.73_{\pm1.41}$ & $79.25_{\pm1.79}$ & $26.33_{\pm4.51}$ \\
         RGCN & $81.85_{\pm3.30}$ & $15.75_{\pm1.67}$ & $69.97_{\pm0.87}$ & $2.77_{\pm0.11}$ & $59.27_{\pm3.93}$ & $18.25_{\pm1.58}$ & $87.59_{\pm1.29}$ & $44.64_{\pm3.67}$ & $61.36_{\pm3.65}$ & $10.68_{\pm1.23}$ \\
         GAT & ${89.52}_{\pm1.18}$ & $33.12_{\pm8.89}$ & $68.68_{\pm1.38}$ & $2.64_{\pm0.16}$ & $55.68_{\pm0.99}$ & $18.31_{\pm0.92}$ & $91.87_{\pm0.74}$ & ${77.30}_{\pm0.92}$ & $83.47_{\pm2.12}$ & $36.86_{\pm7.01}$ \\
         SAGE & $81.58_{\pm1.95}$ & $17.65_{\pm7.12}$ & $73.09_{\pm0.57}$ & $3.06_{\pm0.12}$ & $72.88_{\pm0.82}$ & $32.64_{\pm1.22}$ & $92.36_{\pm1.01}$ & $76.90_{\pm1.42}$ & $88.03_{\pm2.51}$ & $56.70_{\pm5.40}$ \\
         GIN & $87.86_{\pm4.97}$ & $21.96_{\pm5.92}$ & $71.13_{\pm0.89}$ & $2.75_{\pm0.16}$ & $73.28_{\pm1.26}$ & $32.78_{\pm2.26}$ & $90.79_{\pm1.04}$ & $56.90_{\pm7.04}$ & $84.61_{\pm2.05}$ & $43.83_{\pm8.63}$ \\
         \hdashline
         XGBGraph & $87.20_{\pm2.38}$ & ${54.90}_{\pm2.31}$ & $67.02_{\pm0.76}$ & $2.30_{\pm0.07}$ & $74.76_{\pm0.95}$ & $36.49_{\pm1.69}$ & ${93.75}_{\pm3.24}$ & ${81.42}_{\pm3.56}$ & ${91.42}_{\pm3.05}$ & ${72.29}_{\pm10.53}$ \\
         CARE-GNN & $69.81_{\pm4.13}$ & $6.30_{\pm1.19}$ & $69.25_{\pm0.59}$ & $2.18_{\pm0.11}$ & $73.70_{\pm2.68}$ & $33.30_{\pm3.25}$ & $89.51_{\pm0.95}$ & $56.81_{\pm5.18}$ & $88.52_{\pm1.73}$ & $49.54_{\pm6.43}$ \\
         DCI & $82.12_{\pm0.86}$ & $13.48_{\pm1.59}$ & $70.00_{\pm0.52}$ & $2.65_{\pm0.07}$ & $65.71_{\pm7.22}$ & $25.41_{\pm6.34}$ & $91.15_{\pm0.63}$ & $73.79_{\pm2.10}$ & $88.75_{\pm1.62}$ & $49.66_{\pm7.95}$ \\
         BWGNN & $86.27_{\pm2.13}$ & $33.73_{\pm3.61}$ & ${75.25}_{\pm0.63}$ & ${3.38}_{\pm0.15}$ & $71.60_{\pm0.87}$ & $30.73_{\pm1.85}$ & $91.99_{\pm0.96}$ & $73.54_{\pm2.30}$ & $88.40_{\pm0.97}$ & $41.04_{\pm6.59}$ \\
         GHRN & $86.26_{\pm2.32}$ & $42.46_{\pm7.11}$ & ${75.30}_{\pm0.42}$ & ${3.40}_{\pm0.10}$ & $71.71_{\pm0.91}$ & $31.17_{\pm1.70}$ & $91.54_{\pm1.04}$ & $71.55_{\pm3.21}$ & $88.37_{\pm1.03}$ & $40.77_{\pm6.53}$ \\
         PC-GNN & $86.27_{\pm2.13}$ & $25.28_{\pm4.26}$ & $70.75_{\pm0.09}$ & $2.48_{\pm0.19}$ & ${77.99}_{\pm0.84}$ & ${39.91}_{\pm1.76}$ & $91.66_{\pm0.79}$ & $71.27_{\pm4.77}$ & $88.08_{\pm2.44}$ & $45.16_{\pm6.97}$ \\
         ConsisGAD & ${94.01}_{\pm0.22}$ & ${46.68}_{\pm0.61}$ & $73.80_{\pm0.22}$ & $3.26_{\pm0.08}$ & ${82.85}_{\pm0.48}$ & ${45.96}_{\pm1.15}$ & $\textbf{95.33}_{\pm0.63}$ & $\textbf{86.33}_{\pm0.65}$ & $\textbf{91.89}_{\pm1.07}$ & $\textbf{80.39}_{\pm1.67}$ \\
         \hline
         CRoC & $\textbf{95.58}_{\pm0.52}$ & $\textbf{64.24}_{\pm6.06}$ & $\textbf{76.62}_{\pm0.54}$ & $\textbf{3.75}_{\pm0.13}$ & $\textbf{83.57}_{\pm0.87}$ & $\textbf{50.15}_{\pm1.98}$ & $92.44_{\pm0.67}$ & $76.48_{\pm2.03}$ & $89.20_{\pm1.30}$ & $52.98_{\pm7.12}$ \\
         \hline
    \end{tabular}
    }
    \label{tab:exp:perf}
\end{table*}

We show the results in Table \ref{tab:exp:perf}.
CRoC outperforms all other methods in the three largest datasets (i.e., T-Soc, DGraph and Yelp) and is comparable to the state-of-the-art in Amazon and T-Fin.
We draw the following conclusions from the experiment observation:

\begin{itemize}
    \item Context refactoring is effective in enhancing the GNN backbone.
    By training on the context-refactored graph, a GNN is forced to be robust against potential node feature camouflage in the graph.
    The relation-aware joint aggregation distinguishes different kinds of relations, which helps detect behaviour camouflage.
    These mechanisms are particularly effective in Yelp and T-Soc: the former is known to have reviews with camouflage, while the latter, sourced from social networks, includes node feature camouflage.
    This enables our model to outperform camouflage-unaware models by a large margin.
    \item Models trained relying on supervised information may fail to learn sufficient knowledge from such limited labeled data.
    For example, CARE-GNN needs labeled nodes to train its similarity measurement, and GHRN requires labels for edge-pruning.
    In contrast, CRoC leverages both limited labeled data and abundant unlabeled data, empowering the model to capture more diverse patterns than other methods.
    \item With limited supervision, other methods may fall into overfitting.
    In contrast, the capability to learn from unlabeled data endows our model with better generalization ability.
    This advantage is particularly pronounced on large-scale graphs, where CRoC can utilize more unlabeled data for training.
    \item Compared with ConsisGAD, which also focuses on limited supervision scenarios, CRoC shows some distinct advantages:
    (1) To incorporate unlabeled nodes for training, ConsisGAD introduces a learnable module for data augmentation.
    In contrast, CRoC's context refactoring mechanism is a parameter-free data augmentation method that adapts to GAD scenarios, resulting in lower computational complexity.
    (2) ConsisGAD uses predicted pseudo-labels as supervision signals for consistency training, which inevitably introduces noise even though they go through a filtering process.
    Different from ConsisGAD, CRoC always uses the given limited labels as the only supervised signal, avoiding being misled by potential camouflage.
    This makes CRoC perform significantly better than ConsisGAD in hard cases like DGraph.
\end{itemize}

We also observe that the performance gap of all models is minor in Amazon and T-Fin.
In some cases, feature-based methods even outperform GNN models, aligning with findings from GADBench \cite{gadbench}.
This suggests that graph structural information may have limited utility for anomaly detection in these datasets.
In other words, node features may play a much more important role in decision-making.
Consequently, these datasets may not be ideal benchmarks for evaluating specialized GAD models.
We may focus on other datasets for more reliable GAD evaluations.

\subsection{Ablation Studies}

\label{sect:exp:abs:module}

\begin{table}[t]
    \centering
    \caption{The AUC of CRoC(GIN) with different variants.}
    \resizebox{\columnwidth}{!}{
    \begin{tabular}{ccc|c|c|c}
        \hline
        RJA & $L_{ctr}$ & $L^{crf}_{ce}$ & Yelp (1\%) & T-Soc (0.01\%) & DGraph (1\%)\\
        \hline
        \xmark & \xmark & \xmark & $73.28_{\pm1.26}$ & $87.86_{\pm4.97}$ & $71.13_{\pm0.89}$\\
        \cdashline{1-6}
        \xmark & \xmark & \cmark & $72.96_{\pm1.39}$ & $93.03_{\pm0.45}$ & $71.73_{\pm0.71}$\\
        \xmark & \cmark & \xmark & $73.89_{\pm1.61}$ & $90.85_{\pm1.99}$ & $71.75_{\pm0.90}$\\
        \xmark & \cmark & \cmark & $75.00_{\pm1.45}$ & $93.11_{\pm0.66}$ & $72.15_{\pm0.73}$\\
        \cdashline{1-6}
        \cmark & \xmark & \xmark & $79.40_{\pm1.69}$ & $82.18_{\pm3.18}$ & $74.36_{\pm0.58}$\\
        \cmark & \xmark & \cmark & $81.90_{\pm0.83}$ & $94.34_{\pm1.15}$ & $75.07_{\pm0.91}$\\
        \cmark & \cmark & \xmark & $82.97_{\pm1.10}$ & $90.54_{\pm2.79}$ & $75.16_{\pm0.73}$\\
        \cdashline{1-6}
        \cmark & \cmark & \cmark & $83.57_{\pm0.87}$ & $95.58_{\pm0.52}$ & $75.42_{\pm0.75}$\\
        \hline
    \end{tabular}
    }
    \label{tab:exp:module}
\end{table}

We further investigate the effectiveness of each proposed module.
We conduct experiments by enabling/ disabling the RJA, $L_{ctr}$ and $L^{crf}_{ce}$, which correspond to the Relation-aware Joint Aggregation, node-wise contrastive learning, and context refactoring.
All other experimental settings are kept the same as above.

Results of CRoC using GIN as the backbone are listed in Table \ref{tab:exp:module}.
More extended results are provided in \ref{app:supp_result:abs}.
The findings indicate that most modules independently enhance the backbone model’s performance.
One exception is RJA in T-Soc, where the performance declines significantly if it is solely equipped.
This is because RJA introduces additional learnable parameters to encode relations, which demands more data for training.
Without context refactoring and contrastive learning, the model can only learn from the set of labeled nodes (and their neighborhood), which are extremely limited in T-Soc and result in overfitting.
When integrating RJA with $L_{ctr}$ or $L^{crf}_{ce}$, the performance is improved significantly for all cases.
In CRoC, context refactoring and the contrastive learning strategy actively incorporate more unlabeled nodes for training.
With abundant data to learn from, we can apply a more sophisticated model for GAD, even if the labels available for supervision are limited.
This unleashes the learning capacity of the model, enabling it to learn more comprehensive embeddings for downstream tasks.
In addition, as shown in Table \ref{tab:exp:module}, combining $L_{ctr}$ and $L^{crf}_{ce}$ usually leads to a better performance than applying them individually.
This fact reflects that the knowledge learned from these two modules can be complementary to each other, making the model more powerful in GAD.



\subsection{Qualitative Analysis}
\label{sect:exp:qua_ana}

\begin{figure}[tb]
    \centering
    \includegraphics[width=1\linewidth]{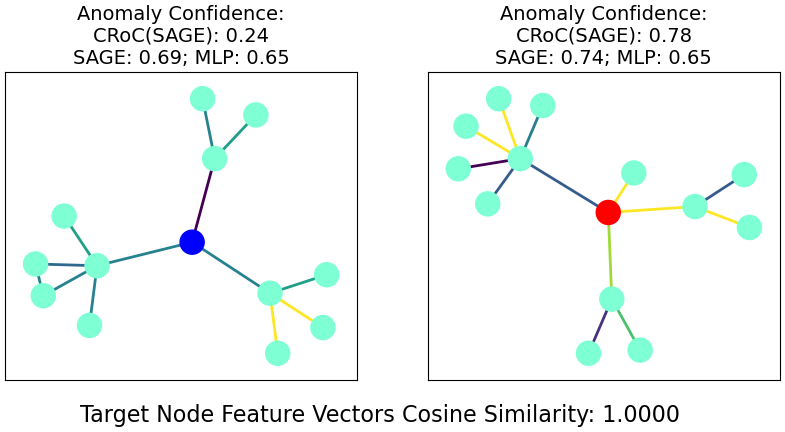}
    \caption{A case study in DGraph: blue (normal) and red (anomalous) nodes are the target nodes. Cyan nodes are neighboring nodes around the target node. Edge color differs for different relations.}
    \label{fig:exp:case_viz}
\end{figure}

We also present some visualization results for qualitative analysis.
Fig. \ref{fig:exp:case_viz} shows a potential camouflage case in DGraph, where anomalous and normal nodes have the same set of node features (cosine similarity is $1$).
In such cases, feature-based algorithms, such as MLP or XGBoost, will definitely fail.
Although common GNNs, such as SAGE \cite{graphsage}, can learn additional information from the neighborhood of the target, they are unaware of the interaction types between nodes.
Consequently, these models struggle to differentiate between two cases with similar local graph structures, just as shown in Fig. \ref{fig:exp:case_viz}.
In contrast, by introducing RJA into the aggregator, the GNN backbone can encode different relations of interactions around the target.
This enables the model to generate a distinct embedding for each node, resulting in better performance in tricky cases, such as behavior camouflage.
More visualization results can be found in \ref{app:supp_result:qua_anal}.


\subsection{Analysis on Computational Efficiency}

CRoC is a plug-and-play scheme to enhance GNN models on GAD tasks, with little additional overhead over the original GNN backbone.
Consider a (message-passing) GNN backbone with $L$ layers, each outputting node embeddings with hidden dimension $F$.
Denote $N$ and $M$ the number of nodes and edges in the graph, respectively.
The overall time complexity of the backbone of a forward step is $O(L(\frac{M}{F} + N)F^{2})$.
After applying CRoC, the time complexity of the corresponding backbone is approximately $O(L(M+N)F^{2})$.
When the graph is sparsely connected (i.e., $M$ and $N$ in the same order, such as DGraph), the CRoC-enhanced GNN maintains the same order of time complexity as the original backbone.
A detailed analysis is provided in \ref{app:discussion:complex_anal}.

\begin{table}[t]
    \centering
    \caption{The averaged time cost (s) of running an update step. GHRN is run on CPU in T-Soc due to GPU out-of-memory.}
    \resizebox{\columnwidth}{!}{
    \begin{tabular}{cccccc}
    \hline
        Dataset & Amazon    & Yelp  & T-Fin & T-Soc & DGraph    \\
        \hline
        GIN     & 0.04      & 0.08  & 0.12  & 0.50  & 0.18      \\
        PCGNN   & 0.03      & 0.03  & 0.05  & 3.02  & 0.61  \\
        BWGNN   & 0.03      & 0.05  & 0.12  & 3.97  & 1.12  \\
        GHRN    & 0.25      & 0.23  & 0.86  & 18.00 (CPU) & 1.42    \\
        ConsisGAD    & 7.29      & 5.88  & 11.47  & 4.21 & 3.52    \\
        \hdashline
        CRoC(GIN) & 0.18    & 0.33  & 0.21  & 1.27  & 0.39  \\
    \hline
    \end{tabular}
    }
    \label{tab:exp:complexity}
\end{table}

We record the time cost of running an update (forward and backward) step of different methods on the same device, as shown in Table \ref{tab:exp:complexity}.
CRoC(GIN) demonstrates linear scalability relative to its backbone (GIN).
Moreover, CRoC(GIN) is significantly faster than other specialized GAD methods in large-scale graphs such as T-Soc and DGraph, highlighting its efficiency advantages.

\subsection{Results on Other Experimental Settings}

\subsubsection{Different Training Label Rate Settings}

\begin{figure}[t]
    \centering
    \includegraphics[width=1\linewidth]{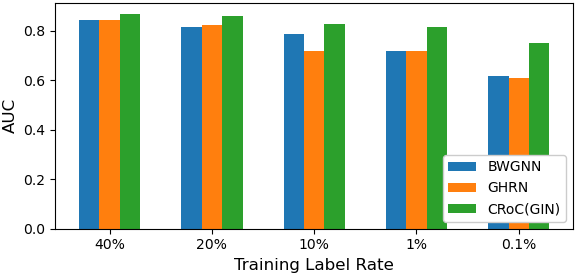}
    \caption{Experimental results for different training label rate settings.}
    \label{fig:exp:diff_tr_rate_yelp}
\end{figure}

We also investigate the performance of CRoC(GIN) when different numbers of labeled samples are available for training.
Experimental results on Yelp are shown in Fig. \ref{fig:exp:diff_tr_rate_yelp}.
We can observe that CRoC(GIN) consistently outperforms other methods.
When fewer training labels are available, the performance of CRoC(GIN) only drops slightly, while other methods suffer from a performance collapse.
Such an observation implies that CRoC can be effective for different GAD scenarios (corresponding to cases where budgets for annotating training data are different).
More experimental results and analyses can be found in \ref{app:supp_result:label_rate}.

\subsubsection{Different Evaluation Protocols}

\begin{table}[t]
    \centering
    \caption{Experimental results (in AUC) on four GAD datasets.}
    \resizebox{\columnwidth}{!}{
    \begin{tabular}{c|cccc}
        \hline
        Dataset & Yelp (1\%) & Amazon (1\%) & Weibo (1\%) & Reddit (1\%) \\
        \hline
        GAD-NR  & $54.36$ & $70.00$ & $87.71$ & $57.99$ \\
        TAM  & $56.43$ & $70.64$ & N/A & $\textbf{60.23}$ \\
        \hdashline
        CRoC(GIN) & $\textbf{81.64}$ & $\textbf{89.65}$ & $\textbf{91.52}$ & $59.92$ \\
         \hline
    \end{tabular}
    }
    \label{tab:exp:diff_proto_unsup}
\end{table}

\begin{table}[t]
    \centering
    \caption{Experimental results (in Macro-F1) on four GAD datasets.}
    \resizebox{\columnwidth}{!}{
    \begin{tabular}{c|cccc}
        \hline
        Dataset & Yelp (1\%) & Amazon (1\%) & T-Fin (1\%) & T-Soc (0.01\%) \\
        \hline
        BSL  & $62.83$ & $\textbf{91.42}$ & $\textbf{86.10}$ & $75.67$ \\
        \hdashline
        CRoC(GIN) & $\textbf{70.87}$ & $74.81$ & $84.26$ & $\textbf{80.80}$ \\
         \hline
    \end{tabular}
    }
    \label{tab:exp:diff_proto_bsl}
\end{table}

There are many specialized GAD models other than the works listed in Table \ref{tab:exp:perf}.
However, due to variations in evaluation protocols (e.g., different datasets, train/test splits, assessment metrics, etc.), it is unable to directly include them \cite{ada-gad,gad-nr,tam,gADAM_cikm,GADAM_iclr,genga,bsl} in Table \ref{tab:exp:perf}.
To ensure fair comparisons, we aligned CRoC's evaluation protocol with their settings and conducted additional experiments.

A part of the results is presented in Tables \ref{tab:exp:diff_proto_unsup} and \ref{tab:exp:diff_proto_bsl}.
We can observe that CRoC(GIN) performs better than other specialized GAD models in most cases.
The advantage of CRoC is more significant when it is applied to large-scale graphs, such as Yelp and T-Soc, which demonstrates the superiority of the proposed framework.
We refer to \ref{app:supp_result:diff_proto} for more detailed comparison results.

\subsection{Extended Experiments and Analyses}


Due to space limitations, other experiments and analyses, such as hyper-parameter investigations, are omitted from the main text.
More explorations and experimental results refer to \ref{app:supp_result}.


\section{Conclusions}

In this paper, we propose Context Refactoring Contrast (CRoC) to enhance GNN models for Graph Anomaly Detection (GAD) tasks.
We start with an analysis of some key challenges of GAD tasks and deal with the problems of camouflage and limited supervision.
Different from previous works, we make use of the class imbalance of GAD and propose context refactoring to augment the original graph.
The objective derived from context refactoring effectively correlates the limited labeled data with the abundant unlabeled data, enabling a GNN to learn more powerful representations by observing more data.
In addition, we propose a relation-aware joint aggregation scheme to enhance the representation power of common GNNs on multi-relation graphs.
By integrating context refactoring with contrastive learning, CRoC-enhanced models can further capture more diverse contextual patterns from the entire graph to compensate for the shortage of supervised signals.
Extensive experiments on different scales of real-world GAD datasets demonstrate the superiority of CRoC over existing works under limited supervision conditions.

\begin{ack}
This work is supported in part by the Innovation and Technology Commission (ITS/244/16) and the CUHK MobiTeC R\&D Fund.
\end{ack}

\appendix
\section{Discussions on Related Works}
\label{app:rel_work}

\subsection{Literature Review of GNNs on GAD}
\label{app:rel_work:gad_review}

GNNs are popular for GAD due to their effectiveness in representation learning on graph-structured data.
However, regarding the challenges specific to GAD, researchers still need to develop proper strategies to adapt GNNs for different scenarios.
For example, the class distribution is typically imbalanced in GAD tasks, where normal nodes account for the majority of the graph.
On the other hand, the minority, i.e., anomalous nodes, may camouflage themselves by connecting to normal nodes.
This raises the concern about neighborhood aggregation with heterophilic connections, which degrades the performance of a typical GNN \cite{ghrn}.
To handle this problem, some works selectively sample nodes for neighborhood aggregation.
Neighbors with higher similarity with the target node are more likely to be sampled.
DAGNN \cite{dagnn} samples Top-K similar nodes from the entire graph for each target.
CARE-GNN \cite{caregnn} introduces a reinforcement learning scheme to adaptively tune the proportion of direct neighbors (of a target node) to be sampled.
PC-GNN \cite{pcgnn} proposes a balanced sampler based on label distributions, where nodes in the minority class have a higher probability of being sampled.

Instead of developing specialized modules to handle the class imbalance problem in GAD, some works utilize this prior condition to train their models.
A typical idea is to learn to reconstruct the graph following an encoder-decoder architecture \cite{gae}.
Since normal nodes account for the majority, the reconstruction process is expected to learn patterns of normal nodes and thereby does well in reconstructing the neighborhood of normal nodes.
In contrast, nodes with bad reconstruction results will be considered anomalous.
To alleviate the impact of anomalous signals, which are regarded as `noise', in training the reconstruction model, ADA-GAD \cite{ada-gad} proposes a series of `denoising' schemes to refine the graph.
It trains multiple encoders on different denoised graphs and introduces an attention mechanism for the decoder.
In this way, the reconstruction error of anomalous nodes is more distinguishable compared to normal nodes.
GAD-NR \cite{gad-nr} opts to reconstruct the entire neighborhood of each node.
To achieve this, GAD-NR proposes a reparameterization method to approximate the neighborhood representation distribution using the multi-variate Gaussian approximation.
Results show that it can outperform many other works when there are some specific types of anomalies in the graph.
Some works utilize the homophily hypothesis or local node affinity assumption (i.e., adjacent nodes should be similar to each other) to train GAD models.
A node with a representation different from the counterpart of its neighbors is considered anomalous.
Along this lane, TAM \cite{tam} proposes to prune non-homophily edges by node features similarity to build multiple new graphs.
A GNN model is trained to maximize the local node affinity on these graphs, which amplifies the differences between the representations of an anomalous node and its neighbors in the original graph.

Another group of works attempts to solve GAD from the perspective of the graph spectrum.
In the work of BWGNN \cite{bwgnn}, the authors observe that the spectral energy of anomalies tends to shift to high-frequency regions.
In contrast, typical GNNs are more likely to be a low-pass filter, which may not work well in such cases.
Therefore, a set of beta-wavelet kernels is combined in parallel to capture the anomalous patterns from different frequencies.
SplitGNN \cite{splitgnn} extends BWGNN by splitting the original graph into a homophilic graph and a heterophilic graph.
Each graph is assigned a set of tunable beta-wavelet filters, adapting to different heterophily levels.

Some other works mine the correlations between graph structures/ node features and anomalous patterns.
GHRN \cite{ghrn} proposes an indicator to measure the heterophily of each edge.
Edge predicted to connect inter-class nodes are pruned to build a new graph.
Such a graph is less heterophilic, so it can be more suitable for neighborhood aggregation for typical GNNs.
GDN \cite{gdn} and GTAN \cite{gtan} focus on the feature space.
The former aims to mine heterophily-invariant features for GNNs, while the latter generates comprehensive embeddings by enriching the node feature source domains.

Self-supervised techniques are also introduced to improve GNNs for GAD.
ANEMONE \cite{anemone} uses a random walk method to generate a local-view augmentation.
A readout function is used to generate a context-view augmentation of a target node.
With different views, the model is trained via contrastive learning.
SLADE \cite{slade} learns to detect edge-stream anomalies by contrasting representations learned from different time scales.
DCI \cite{dci} extends DGI \cite{dgi} by introducing a cluster-based global view for contrastive pre-training.
BSL \cite{bsl}, gADAM \cite{gADAM_cikm}, and ConsisGAD \cite{consisgad} develop specific schemes to build augmentations for node representations and apply the consistency constraint to train their model.

However, the drawbacks of previous works are also obvious:
\begin{enumerate}
    \item When training GAD models based on some inductive biases (e.g., the class imbalance prior condition or the graph homophily hypothesis), there is a tradeoff between the learning capability of the backbone model and the performance.
    For example, reconstruction-based GAD models should not be powerful enough to overfit anomalous instances.
    This leaves an obstacle for applying more advanced GNN models to solve GAD problems.
    In addition, many of this kind of model \cite{tam,gad-nr,ada-gad} are insensitive to camouflage, which limits their generalization to different GAD scenarios.
    \item When dealing with camouflage, the actions of existing works aiming at detecting these camouflaged nodes directly \cite{bwgnn,splitgnn} or indirectly \cite{caregnn,dagnn}.
    The power of their model (or other modules of their model) relies heavily on the performance of these detectors, which are hard to tune or train and sometimes even vulnerable.
    \item Many aforementioned works require abundant labeled anomalous data to train their GNN backbones or related modules, e.g., the node-wise similarity predictor of CARE-GNN or the edge heterophily indicator in GHRN.
    The effectiveness of these modules can be challenged when labels are very limited in the graph, which is common in real-world GAD scenarios. 
    \item When working on multi-relation graphs, existing works either apply an `intra-relation then inter-relation' aggregation scheme in each layer, or split the original multi-relation graph into multiple single-relation graphs to adapt to homogenous graph aggregation.
    Such operations physically decouple different relations, making it hard to learn the correlation among different types of relations.
    When handling some cases like money laundering, where multiple relations (types of payment) are involved simultaneously, the model may fail to perceive anomalous patterns that should be learned across different relations.
\end{enumerate}

\subsection{Comparison with Related Works}
\label{app:rel_work:comp_rel_work}

Our work is partly motivated by some previous methods on semi-supervised learning and data augmentations, including Mixup \cite{mixup}, DCI \cite{dci}, and CoCoS \cite{cocos}, and relation-aware learning methods.

\subsubsection{Comparison with Mix-up}
\label{app:rel_work:comp_rel_work:mixup}
Mixup is a well-recognized scheme for data augmentation, which generates new training sample-label pairs by proportionally mixing two different samples' raw features and their ground-truth labels.
In other words, Mix-up assumes that a convex combination of two samples in the data space also leads to a linear interpolation in the target (label) space.
In this work, we propose context refactoring (refer to Equ \ref{equ:method:feat_mix} in the main text) following the style of Mixup to recompose node features for the context-refactored graph.
However, our scheme randomly samples a node from the entire graph (dataset) to mix with the target node, regardless of whether it is labeled or not, while Mixup only mixes two instances collected from two different classes in the training set (labeled instances).
Since the training set is a subset of the entire graph, the convex hull of the generated samples formed by the original Mixup is definitely a subspace of that of context refactoring.
Therefore, models trained on the context-refactored graph are expected to learn more diverse patterns than those of Mixup.
This can be more effective when labeled data is very limited, such as GAD scenarios.
In addition, we make use of the class-imbalanced prior of GAD to implement context refactoring, which is supposed to keep the context of each node consistent.
Concretely, we preserve the same set of labels (if available) for the nodes in the context-refactored graph.
In contrast, Mixup interpolates the labels of the two selected samples for the newly generated sample.
This step introduces ambiguity for classification tasks and may not work in some special cases. 
Therefore, compared with Mixup, a CRoC-enhanced model can be trained towards a clearer direction supervised by a small set of unambiguous ground-truth labels, resulting in more discriminative representations for GAD tasks.
Quantitative comparison between Mixup and context refactoring is shown in Table \ref{tab:supp_exp:abs_gin} and Table \ref{tab:supp_exp:abs_sage}.

\subsubsection{Comparison with Relation-aware Learning Methods}
\label{app:rel_work:comp_rel_work:ral_method}

Some GNN-based methods attempt to encode relation information when learning node embeddings.

RGCN \cite{rgcn} and its follow-up CARE-GNN \cite{caregnn} both follow a two-step hierarchical aggregation paradigm to learn relation-related information.
In each GNN layer, the aggregator first aggregates information from neighbors that keep the same kind of relation with the target node.
This step generates a `representative' embedding of each type of relation for a target node.
In the second step, the aggregator reduces embeddings from multiple relations into a single representation for the target node in the current layer.

Another way is to decouple the multi-relation graph into multiple single-relation graphs, such as BWGNN \cite{bwgnn} and SplitGNN \cite{splitgnn}.
GNNs are applied to each single-relation graph to generate relation-specific embeddings for each node for each relation type.
The embeddings of the same node with respect to different relations are finally combined (e.g., through concatenation) to yield a comprehensive representation.

Obviously, most previous works deliberately decouple the information of different relations during the aggregation process.
However, connections in the graph with different relations are not absolutely independent of each other.
For example, a money launderer is very likely to transfer a large amount of money through different payment platforms (relations), while an illicit payment record in a platform may hide in many valid payment records to camouflage the malicious activity.
Therefore, previous works that encode relations may not capture fine-grained correlations between two nodes involved in different relations, which may lose track of some anomalous cases with complicated patterns.

In contrast, in the proposed Relation-aware Joint Aggregation (RJA) module, we explicitly encode each kind of relation and directly associate a relation with the neighbor who will interact with the target node.
This enables a GNN to utilize the aggregated information across different relations, encouraging it to mine the mutual correlations among different types of interactions.
This design benefits a GNN in distinguishing anomalous instances with complicated interaction or behavior patterns.
Quantitative results (refer to \ref{app:supp_result:abs}) and qualitative results (refer to \ref{app:supp_result:qua_anal}) both demonstrate the effectiveness of RJA.

\subsubsection{Comparison with DGI}
\label{app:rel_work:comp_rel_work:dgi}

DCI decouples the learning process for GAD models by a two-stage paradigm: it first pretrains the model in a self-supervised manner and then fine-tunes all learnable parameters in a supervised manner using the labeled data.
In the pretraining stage, K-means is applied to cluster each node.
The cluster centroid will be paired with a node in the same cluster to be a positive sample for contrastive learning.
By applying such a two-stage training scheme, DCI is able to learn from both labeled and unlabeled data.
However, since it is task-agnostic during the pretraining stage, the learned knowledge is not guaranteed to be compatible with/ informative to the downstream task.
Although DCI will be fine-tuned on the labeled set after pretraining, the model may suffer from unlearning due to the nature of the decoupled training paradigm, especially when labeled set is very small.
In contrast, CRoC unifies the supervised and self-supervised learning objectives to jointly train the model in an end-to-end manner.
This encourages the model to learn complementary knowledge from different objectives, which improves the representation ability of a model.

\subsubsection{Comparison with CoCoS}
\label{app:rel_work:comp_rel_work:cocos}

CoCoS proposes a context-sharing method to exchange node features within the same class based on the predicted pseudo-labels for common node classification tasks.
Pseudo-labeling is an effective way to improve data utilization.
However, this kind of method can hardly work on GAD tasks due to class imbalance of GAD data.
Instead, CRoC makes use of the class imbalance prior of GAD for context refactoring, which builds a graph with similar semantics to the original graph without using any supervised information or estimated labels.
A CRoC-enhanced model can, therefore, avoid the risk of error accumulation induced by pseudo-labeling, which is more suitable for GAD tasks.
In addition, although CoCoS and CRoC both introduce a contrastive learning regularization term to learn from unlabeled nodes, the former uses an additional classifier to distinguish positive and negative samples.
On the contrary, CRoC incorporates the self-supervised learning objective by using the InfoNCE loss without introducing any additional learnable parameters.
This makes CRoC a more lightweight and plug-and-play framework to enhance GNN-based models on GAD tasks, which is more effective and efficient.


\section{Extended Discussions on CRoC}
\label{app:discussion}

\subsection{Context Refactoring for Correlating Labeled and Unlabeled Data}
\label{app:discussion:lb_ulb}

In CRoC, the context refactoring strategy makes use of the class imbalance of the GAD task to build a refactored graph through:
\begin{equation}
    \bm{\tilde{X}} =  \Omega(\bm{X})
    \label{equ:appendix:feat_shuf}
\end{equation}
\begin{equation}
    \bm{X'} = \alpha\bm{X} + (1 - \alpha)\bm{\tilde{X}}
    \label{equ:appendix:feat_mix}
\end{equation}
$\bm{X'}$ is the refactored node feature matrix, and the context-refactored graph for each training epoch is denoted as $\mathcal{G}' = (\mathcal{V}, \bm{X'}, \{\mathcal{E}_{r}\}|_{r=1}^{R})$.
The GNN backbone is then trained jointly with the objective $L^{crf}_{ce}$, which is defined as:
\begin{equation}
    \label{equ:appendix:crf_loss}
    L^{crf}_{ce} = -\frac{1}{|\mathcal{V}^{L}|} \sum_{v\in\mathcal{V}^{L}}\bm{y}_{v}^{\intercal}\log{(\hat{\bm{y}}'_{v})}
\end{equation}
where $\hat{\bm{y}}'_{v}$ is the prediction of node $v$ in the context-refactored graph yielded by the GNN backbone.

Recall that GNN generates an embedding for a target node by aggregating the information from its neighbors.
The objective $L^{crf}_{ce}$ is thereby a function of node features in the graph.
Let's denote $N(\mathcal{V}^{L})$ as the set of unlabeled nodes in the neighborhood of labeled nodes and $\bm{X}_{Q}$ as the node feature matrix corresponding to the node set $Q$.
Then we can rewrite $L^{crf}_{ce}$ as:
\begin{align}
    L^{crf}_{ce} &= f(\bm{X'}_{\mathcal{V}^{L}},\bm{X'}_{N(\mathcal{V}^{L})}) \\
                 &= f(\bm{X}_{\mathcal{V}^{L}},\bm{X}_{N(\mathcal{V}^{L})}, \bm{\tilde{X}}_{\mathcal{V}^{L}}, \bm{\tilde{X}}_{N(\mathcal{V}^{L})})
\end{align}
For comparison, the original supervised objective can also be rewritten as:
\begin{equation}
    L^{ori}_{ce} = f(\bm{X}_{\mathcal{V}^{L}},\bm{X}_{N(\mathcal{V}^{L})})
\end{equation}
Note that unlabeled nodes are the majority of the graph and $|N(\mathcal{V}^{L})| << |\mathcal{V}^{U}|$, the node features being shuffled to $\bm{\tilde{X}}_{N(\mathcal{V}^{L})}$ are with high probability to be originated from unlabeled nodes that located outside the neighborhood of any labeled node.
Therefore, when optimizing the GNN by the objective $L^{crf}_{ce}$, we are essentially encouraging the model to explore the correlation between labeled data ($\bm{X}_{\mathcal{V}^{L}}$) and the extra unlabeled data ($\bm{\tilde{X}}_{\mathcal{V}^{L}}$ and $\bm{\tilde{X}}_{N(\mathcal{V}^{L})}$).
Moreover, the node feature matrix $\bm{X}_{\mathcal{V}^{L}}$ and $\bm{X}_{N(\mathcal{V}^{L})}$ are fixed for the entire training process.
In contrast, due to the randomness raised by the random shuffle function $\Omega(\cdot)$, $\bm{\tilde{X}}_{N(\mathcal{V}^{L})}$ will be reassigned in each epoch.
By training the GNN backbone for multiple epochs, we can actively incorporate more unlabeled data into the training process.
Although there can be undesired shuffling results for some of the epochs (e.g., a normal node is assigned the features from an anomalous node), these cases can be extremely few because of the class imbalance prior of GAD, which causes little effect for the entire training process.
Compared with the original supervised objective $L^{ori}_{ce}$ and the Mix-up scheme \cite{mixup}, this strategy effectively extends the input feature space of the GNN model, which prevents the model from overfitting the data only related to the few labeled nodes (i.e., $\bm{X}_{\mathcal{V}^{L}}$ and $\bm{X}_{N(\mathcal{V}^{L})}$).
In addition, different from some methods that use pseudo-labels to train the model \cite{adaedge,m3s}, $L^{crf}_{ce}$ is fully supervised by the ground-truth labels.
This avoids the risk of learning from incorrect knowledge and thereby generates more discriminative embeddings.

\subsection{Context Refactoring Contrast for Learning More Diverse Patterns}
\label{app:discussion:diverse_pattern}

In CRoC, we integrate context refactoring with a contrastive learning paradigm to enhance the backbone GNN.
This enables the model to learn from unlabeled data, which encourages it to capture more diverse patterns from the graph.

As shown in Equ (\ref{equ:method:feat_mix}), the node feature matrix for the context-refactored graph ($\bm{X'}$) is a convex combination of the original node feature matrix ($\bm{X}$) and the shuffled node feature matrix ($\bm{\tilde{X}}$).
By tuning the hyper-parameter $\alpha$, we can control the magnitude to preserve the native node features for context refactoring.
This enables us to generate some intermediate samples that can make the input feature space smoother or contiguous. 
Moreover, $\bm{X'}$ can be a set of new features with some of the element values never shown in $\bm{X}$.
Training the GNN backbone on the context-refactored graph essentially forces the model to adapt to different kinds of feature combinations, which encourages it to learn more diverse patterns that can be beneficial for making decisions.

From the perspective of graph topology, the model enhanced by CRoC can observe more different types of local graph structures than the original backbone.
This is owed to the introduction of the contrastive learning paradigm, endowing it with the capability to learn from all unlabeled nodes.
Denote the induced computational graph of the GNN backbone in node $v$ as $\mathcal{G}_{v}$.
Note that the original GNN is trained only by optimizing the supervised objective $L^{ori}_{ce}$.
Then, the set of induced computational graphs fed to train the original GNN is $S_{\mathcal{V}^{L}} = \{ \mathcal{G}_{v} | v \in \mathcal{V}^{L} \}$.
In contrast, by introducing contrastive learning, the CRoC-enhanced model actively incorporates all unlabeled nodes into the training process.
In this way, the set of induced computational graphs fed to train the CRoC-enhanced model is $S_{\mathcal{V}} = \{ \mathcal{G}_{v} | v \in \mathcal{V} \}$.
Since $\mathcal{V}^{L} \subseteq \mathcal{V}$, it is always true that $S_{\mathcal{V}^{L}} \subseteq S_{\mathcal{V}}$.
Therefore, the local graph structures that the CRoC-enhanced model can observe are more diverse than those of the vanilla model during training.
This endows the CRoC-enhanced model with the ability to learn more diverse structure-related patterns for generating more discriminative representations for the GAD task.

\subsection{Complexity Analysis}
\label{app:discussion:complex_anal}

We begin with the analysis of the computation complexity of a common message-passing GNN model, such as GraphSAGE \cite{graphsage}.
Suppose the GNN backbone stacks $L$ layers, each of which outputs node embeddings with hidden dimension $F$.
Denote $N$ and $M$ the number of nodes and edges in the graph, respectively.
Then, the time complexity of message propagation (corresponding to aggregators) in the entire graph is $O(LMF)$.
Accordingly, the time complexity of update functions, which are typically linear transformation operations, is $O(LNF^{2})$.
Then the overall time complexity for a common GNN model on a forward step is:
\begin{equation}
    T_{GNN} = O(LMF + LNF^{2})
\end{equation}

When enhancing the backbone GNN with CRoC, the computational overhead is mainly raised by the new proposed modules and strategies, i.e., the context refactoring operation, Relation-aware Joint Aggregation (RJA), and the contrastive learning paradigm.
Note that context refactoring and the contrastive learning scheme are only implemented for training. 
The following analyses are for the training stage. 
For context refactoring, the main operations are random feature shuffling and convex combinations (refer to Equ (\ref{equ:method:feat_shuf}) and (\ref{equ:method:feat_mix}) in the main text), but they are conducted once for each training epoch at the very beginning.
Therefore, these two operations take time with complexity $O(NF)$.
RJA is implemented together with the message propagation mechanism.
The extra computation comes from Equ (\ref{equ:method:edge_emb}) in the main text, where it takes the relation embedding of a specific edge as input and fuses it with the source node embedding.
The computation complexity of this part is $(F+F)\times F \times M \times L=O(LMF^{2})$.
Other operations of RJA are the same as those of common message propagation.
As for the contrastive learning paradigm, suppose we sample $k$ negative samples to train with each positive sample.
According to Equ (\ref{equ:method:ctr_loss}) in the main text, the time complexity for contrastive learning will be $O((k+1)NF)$.
Therefore, the overall time complexity of a CRoC-enhanced GNN is:
\begin{equation}
\begin{split}
      T_{CRoC} = O(NF) + O(LMF+LNF^{2}+LMF^{2})\\
              + O(LMF+LNF^{2}+LMF^{2}) + O((k+1)NF)               
\end{split}
\end{equation}
where the second and third terms are the time cost of a forward step in the original graph and the context-refactored graph by a GNN with RJA. 
We can further simplify the time complexity as follows:
\begin{equation}
     T_{CRoC} \approx O(L(M+N)F^{2})
\end{equation}
In sparsely connected graphs ($M$ and $N$ in the same order, e.g., DGraph), the time complexity of a CRoC-enhanced GNN can be linear to the vanilla model. 
For dense graphs ($M >> N$, e.g., T-Fin), the additional overhead correlates with the hidden dimension $F$.

In the inference stage, no context refactoring and contrastive learning operations are applied.
Therefore, the inference time complexity of a CRoC-enhanced GNN is on par with its vanilla model.

As for the space complexity, we only introduce additional learnable parameters in RJA (for encoding the relation embedding and the linear transformation for fusing the relation embedding and the source node embedding).
In the training stage, CRoC is required to generate node embeddings for both the original graph and the context-refactored graph for each forward step.
Therefore, the memory consumption of a CRoC-enhanced GNN will roughly be twice that of the vanilla model.
This may become a bottleneck when applying full-batch training on a large-scale or densely connected graph.
To address this problem, in our experiments, we apply random neighbor-sampling \cite{graphsage} for training, which essentially trains the model on a set of subgraphs sampled from the entire graph.
This trick significantly reduces the memory demand as well as the time complexity of CRoC for each training step.

\begin{table}[t]
    \centering
    \caption{The averaged runtime (s) for taking an update step (forward and backward). (In T-Soc, GHRN is run in CPU due to the GPU out-of-memory problem.)}
    \resizebox{\columnwidth}{!}{
    \begin{tabular}{cccccc}
    \hline
        Dataset & Amazon    & Yelp  & T-Fin & T-Soc & DGraph    \\
        \hline
        GIN     & 0.04      & 0.08  & 0.12  & 0.50  & 0.18      \\
        PCGNN   & 0.03      & 0.03  & 0.05  & 3.02  & 0.61  \\
        BWGNN   & 0.03      & 0.05  & 0.12  & 3.97  & 1.12  \\
        GHRN    & 0.25      & 0.23  & 0.86  & 18.00 (CPU) & 1.42    \\
        ConsisGAD    & 7.29      & 5.88  & 11.47  & 4.21 & 3.52    \\
        \hdashline
        CRoC(GIN) & 0.18    & 0.33  & 0.21  & 1.27  & 0.39  \\
    \hline
    \end{tabular}
    }
    \label{tab:discuss:complexity}
\end{table}

We record the averaged runtime of CRoC(GIN) for taking an update step (i.e., a forward and backward step) during training.
The time cost of other models (implemented by the GADBench \cite{gadbench}) under the same hardware environment is also provided for comparison.
The results are listed in Table \ref{tab:discuss:complexity}.
It can be observed that the time cost of CRoC(GIN) is linear to its vanilla model, i.e., GIN, in small or middle-scale datasets (Amazon, Yelp, and T-Fin) and on par with other methods.
However, when the input graph becomes larger, such as T-Soc and DGraph, CRoC(GIN) runs much faster than other cutting-edge GAD models.
These empirical results demonstrate that the computation efficiency is comparable to other models in small or medium-sized graphs and more efficient than other state-of-the-art methods in large-scale graphs.

\section{Details of Experiment Settings}

\label{app:exp_setting}



\subsection{Datasets}
\label{app:exp_setting:dataset}

\begin{table*}[t]
    \centering
    \caption{Graph Anomaly Detection Datasets Statistics.}
    \begin{tabular}{c|ccccc}
        \hline
         Dataset    &  \#Node   &  \#Edge       &   \#Anomaly   &   Anomaly(\%) & Density\\
         \hline
         Weibo     &  8,405   &  407,963    &   868         &   10.33\%   & $1.16\times 10^{-2}$ \\
         Reddit     &  10,984   &  168,016    &   366         &   3.33\%   & $0.28\times 10^{-2}$ \\
         Amazon     &  11,944   &  9,557,648    &   821         &   6.87\%   & $6.7\times 10^{-2}$ \\
         Yelp       &  45,954   &  8,051,348    &   6,677       &   14.53\%   & $0.38\times 10^{-2}$  \\
         T-Fin      &  39,357   &  21,222,543   &   1,803       &   4.58\%   & $2.74\times 10^{-2}$   \\
         T-Soc      &  5,781,065   &  73,105,508   &   174,280     &   3.01\%    & $4.37\times 10^{-6}$  \\
         DGraph     &  3,700,550   &  4,300,999    &   15,509      &   0.42\%    & $6.28\times 10^{-7}$  \\
         \hline
    \end{tabular}
    \label{tab:exp_setting:ds_stat}
\end{table*}

In our experiments, we evaluate CRoC on five real-world datasets that are publicly accessible.
The statistics of each dataset are provided in Table \ref{tab:exp_setting:ds_stat}.

\subsubsection{YelpChi} 

YelpChi (Yelp in short) \cite{yelp} is collected from yelp.com for spam review detection.
Each node in the graph is a user review, either for hotels or restaurants, filtered (spam) or recommended (legitimate) by Yelp.
Nodes can be connected based on pre-defined relationships, and there are three different relations in the graph.
In this dataset, each review (node) is represented by a 32-dim feature vector.

\subsubsection{Amazon} 
This dataset \cite{amazon} is collected from Amazon's product reviews to detect users who write fake reviews.
Each node is a user, and edges associated with different relations can be built between two users using handcrafted rules.
Each node is represented by a 25-dimensional vector, and three relations are included in the graph.

\subsubsection{T-Social} 
T-Social (T-Soc in short) \cite{bwgnn} aims to find anomalous accounts in a social network, where nodes are accounts, and an edge indicates two accounts hold a friendship relation.
Each node has a 10-dim feature vector, and only one edge relation is provided.

\subsubsection{T-Finance} 
T-Fin \cite{bwgnn} aims to find anomalous accounts in a transaction network.
Each edge connects two accounts if they have transaction records.
Other settings are the same as T-Soc.

\subsubsection{DGraph-Fin} 
DGraph-Fin (DGraph in short) \cite{dgraph} aims to detect fraudsters in a financial network.
Each node corresponds to a user, while an edge between two users indicates one of the 19 contact relations of the graph.
In this dataset, each node is represented by a 17-dimensional feature vector, which is derived from the basic personal profile of a user.
However, some elements of the node feature vector are missing.
These missing features are consistently padded by the value `$-1$' by default.

\subsubsection{Weibo}
Weibo \cite{weibo} is a user-user interaction dataset collected from Tencent-Weibo.
Each node corresponds to a user in the graph, and a 300-dim bag-of-words vector is used to represent the node features.
A user is supposed to be anomalous if he/she is found conducted at least five suspicious events within a certain range of time, otherwise benign.

\subsubsection{Reddit}
Reddit \cite{reddit} is a user-subreddit interaction dataset collected from the Reddit platform.
The text contents of users' posts and subreddits are converted into a feature vector representing their LIWC categories.
The summation of these vectors is used to represent the node features of each user or subreddit.
A user node is labeled as anomalous if he/she is banned from a subreddit.

\subsection{Implementation Details}
\label{app:exp_setting:imp_detail}

\subsubsection{Model Implementation}

We mainly compare CRoC with three different kinds of methods, including:
\begin{enumerate}
    \item Common classifiers: MLP, XGBoost (XGB). These methods make decisions only based on features without considering graph structures.
    \item GNN baselines: GCN, GAT, GraphSAGE (SAGE), GIN. These methods are typical models of semi-supervised learning for graph-structured data.
    \item GAD-specific models: CAREGNN, DCI, PCGNN, BWGNN, XGBGraph, and GHRN. These models are cutting-edge methods specialized to address different kinds of GAD tasks.
\end{enumerate}

In our experiments, we implement XGBoost, XGBGraph, CARE-GNN, PCGNN, DCI, BWGNN, and GHRN using the GADBench toolbox \cite{gadbench}.
Others are all implemented based on the DGL library.

As for our proposed method, CRoC, we implement it based on two GNN backbones: GraphSAGE and GIN.
In general, we stack two layers for the GNN backbone, which is followed by a two-layer MLP as the classifier.
An exception is the implementation in the smallest dataset, Amazon, where predictions are yielded directly by the GNN backbone.
To handle large-scale graphs, we apply neighbor-sampling strategies following the work of GraphSAGE \cite{graphsage} for mini-batch training. 
Since we stack two layers for our GNN backbone, we randomly sample ten neighbors in the first aggregation layer and five neighbors for the second layer.
To deal with the class-imbalanced problem, we downsample labeled normal nodes to train the model.
(The same training strategy is applied to other baselines for fair comparison.)
For more details on the implementation of our proposed model, please refer to the code provided in the supplementary materials.

\subsubsection{Hyper-parameter Setting/ Tuning}


In Equ (\ref{equ:method:ctr_loss}) of the main text, we fix the temperature as $\tau=2$ and let $k=10$, i.e., sample 10 negative samples for each positive sample.

In our experiments, there are three tunable coefficients, including $\alpha$, $\gamma$, and $\eta$.
For $\alpha$, we search it in $\{ 0, 0.2, 0.25, 0.5, 0.75, 0.9\}$.
For $\gamma$ and $\eta$, we searched them in $\{ 0.2, 0.4, 0.5, 0.6, 0.8, 1\}$.

In our experiments, we use Adam to optimize the learnable parameters.
The learning rate is $0.003$.

\subsubsection{Experimental Setup}

We follow the settings of BWGNN \cite{bwgnn} to split the datasets, where we randomly sample $1\%$ ($0.01\%$ for T-Soc) of nodes as labeled nodes to form the training set.
The remaining one-third and the other two-thirds serve as the validation and testing set.

We measure the performance using the Area under the ROC Curve (AUC) and the Averaged Precision score (AP).
In our experiment, we use ten different random seeds to initiate the dataset split for the training and evaluation of each model.
We run each model based on these ten random seeds, and we record the testing results with respect to the best validation results for each run.
Finally, we average the results over all ten rounds and report the number in this paper.

\subsection{Configurations for Result Reproducing}
\label{app:exp_setting:config}

\subsubsection{Hyper-paraters}


\begin{table*}[t]
    \centering
    \caption{Detailed hyper-parameter settings for reproducing results.}
    \begin{tabular}{c|c|c|c|c|c|c|c}
        \hline
        Coefficient & Weibo & Reddit & Yelp & Amazon & T-Soc & T-Fin & DGraph \\
        \hline
        $\alpha$ & 0.5 & 0 & 0 & 0.5 & 0 & 0.5 & 0.5 \\
        $\gamma$ & 0.5 & 0.5 & 0.2 & 0.2 & 1 & 0.2 & 1 \\
        $\eta$  & 0.5 & 0.5 & 0.5 & 0.5 & 0.5 & 0.2 & 0.5 \\
        hid dim & 64 & 64 & 64 & 64 & 64 & 64 & 64 \\
        \#epoch & 200 & 200 & 300 & 200 & 200 & 200 & 150 \\
        \hline
    \end{tabular}
    \label{tab:supp_exp:hp_details}
\end{table*}

We tune the hyper-parameters for our model following the strategies mentioned above.
Final hyper-parameter settings for reproducing the reported results on each dataset can be found in Table \ref{tab:supp_exp:hp_details} in this supplementary material.


\subsubsection{Hardware Configurations}

Our experiments are conducted on a single Linux machine with an AMD Ryzen Threadripper 3970X CPU
@ 3.7GHz, 256 GB RAM, and an NVIDIA RTX TITAN GPU with 24 GB RAM.

\section{Supplementary Experimental Results}
\label{app:supp_result}

\subsection{Experimental Results of CRoC Variants}
\label{app:supp_result:croc_perf}



\begin{table*}
    \centering
    \caption{Experimental Results on Five GAD Datasets. (The label ratio of each dataset is marked in parentheses.)}
    \resizebox{2.1\columnwidth}{!}{
    \begin{tabular}{c|cc|cc|cc|cc|cc}
        \hline
         Dataset & \multicolumn{2}{c|}{Yelp (1\%)} & \multicolumn{2}{c|}{Amazon (1\%)} & \multicolumn{2}{c|}{T-Soc (0.01\%)} & \multicolumn{2}{c|}{T-Fin (1\%)} & \multicolumn{2}{c}{DGraph (1\%)} \\
         \hline
         CRoC (SAGE)& $79.60_{\pm0.74}$ & $40.79_{\pm2.15}$ & $89.20_{\pm1.30}$ & $52.98_{\pm7.12}$ & $88.79_{\pm1.35}$ & $26.27_{\pm6.69}$ & $92.44_{\pm0.67}$ & $76.48_{\pm2.03}$ & $76.62_{\pm0.54}$ & $3.75_{\pm0.13}$ \\
         CRoC (GIN)& $81.64_{\pm0.83}$ & $45.87_{\pm2.30}$ & $89.65_{\pm2.15}$ & $56.77_{\pm7.42}$ & $95.58_{\pm0.52}$ & $64.24_{\pm6.06}$ & $91.21_{\pm1.43}$ & $55.09_{\pm11.16}$ & $75.42_{\pm0.75}$ & $3.54_{\pm0.16}$ \\
         \hline
    \end{tabular}
    }
    \label{tab:supp_exp:croc_perf}
\end{table*}

In Table \ref{tab:supp_exp:croc_perf}, we list the detailed experimental results of CRoC with respect to SAGE and GIN.
Compared with the results reported in Table 3 in our main text, both these two CRoC variants outperform most competitive methods.

\subsection{CRoC for Different GNN Backbones}
\label{app:supp_result:croc_backbone}

\begin{table}[t]
    \centering
    \caption{The AUC of Different GNN Backbones with CRoC (w/o RJA). The improvement over the original backbone is attached in parentheses.}
    \begin{tabular}{c|c|c|c}
        \hline
        Model & Yelp (1\%) & T-Soc (0.01\%) & DGraph (1\%) \\
        \hline
        GCN & $53.99(-1.82)$ & $89.66(+7.81)$ & $69.86(+0.61)$ \\
        GAT & $53.92(-1.76)$ & $89.67(+0.15)$ & $69.26(+0.58)$ \\
        SAGE & $73.94(+1.06)$ & $88.54(+6.96)$ & $73.45(+0.36)$\\
        GIN & $75.00(+1.72)$ & $93.11(+5.25)$ & $72.27(+1.14)$ \\
         \hline
    \end{tabular}
    \label{tab:supp_exp:croc_backbone}
\end{table}

As a plug-and-play scheme, we further evaluate CRoC by using different kinds of GNN backbones.
For fair comparisons, we do not implement the Relation-aware Joint Aggregation (RJA), which introduces additional learnable parameters, since our target is to investigate the impact of CRoC on them.

Our experiments are conducted on Yelp, T-Soc and DGraph with the same settings above.
Results are shown in Table \ref{tab:supp_exp:croc_backbone}.
Obviously, almost all listed GNN backbones get a performance gain by training with CRoC.
GIN even gains a $11\%$ improvement on Yelp.
The two exceptions are GCN and GAT on the Yelp dataset.
It is speculated that GCN and GAT are unable to detect anomalies in Yelp since their vanilla model also performs badly in the same case.
This may be because target node features in Yelp can be informative, while GCN and GAT mix them with the features collected from their neighbours.
Instead, SAGE and GIN both keep a bypass from the target node to differentiate from neighbors, making them survive in Yelp.

It is worth mentioning that CRoC can be more helpful when the context information of the graph is important for detection, e.g., in T-Soc.
When only very limited labels are provided for training, CRoC can drive the model to learn from the context of unlabeled nodes.
This significantly enhances a GNN even under limited supervision, which results in better performance gain.

\subsection{Supplementary Results of Ablation Studies}
\label{app:supp_result:abs}



\begin{table}[t]
    \centering
    \caption{The AUC of CRoC (GIN) with Different Variants.}
    \resizebox{\columnwidth}{!}{
    \begin{tabular}{ccc|c|c|c}
        \hline
        RJA & $L_{ctr}$ & $L^{crf}_{ce}$ & Yelp (1\%) & T-Soc (0.01\%) & DGraph (1\%)\\
        \hline
        \xmark & \xmark & \xmark & $73.28_{\pm1.26}$ & $87.86_{\pm4.97}$ & $71.13_{\pm0.89}$\\
        \cdashline{1-6}
        \xmark & \xmark & Mixup & $63.44_{\pm3.07}$ & $75.64_{\pm12.58}$ & $75.00_{\pm0.51}$\\
        \xmark & \xmark & \cmark & $72.96_{\pm1.39}$ & $93.03_{\pm0.45}$ & $71.73_{\pm0.71}$\\
        \xmark & \cmark & \xmark & $73.89_{\pm1.61}$ & $90.85_{\pm1.99}$ & $71.75_{\pm0.90}$\\
        \xmark & \cmark & \cmark & $75.00_{\pm1.45}$ & $93.11_{\pm0.66}$ & $72.15_{\pm0.73}$\\
        \cdashline{1-6}
        \cmark & \xmark & \xmark & $79.40_{\pm1.69}$ & $82.18_{\pm3.18}$ & $74.36_{\pm0.58}$\\
        \cmark & \xmark & Mixup & $80.54_{\pm0.75}$ & $74.23_{\pm5.60}$ & $75.00_{\pm0.51}$\\
        \cmark & \xmark & \cmark & $81.90_{\pm0.83}$ & $94.34_{\pm1.15}$ & $75.07_{\pm0.91}$\\
        \cmark & \cmark & \xmark & $82.97_{\pm1.10}$ & $90.54_{\pm2.79}$ & $75.16_{\pm0.73}$\\
        \cdashline{1-6}
        \cmark & \cmark & \cmark & $83.57_{\pm0.87}$ & $95.58_{\pm0.52}$ & $75.42_{\pm0.75}$\\
        \hline
    \end{tabular}
    }
    \label{tab:supp_exp:abs_gin}
\end{table}

\begin{table}[t]
    \centering
    \caption{The AUC of CRoC (SAGE) with Different Variants.}
    \resizebox{\columnwidth}{!}{
    \begin{tabular}{ccc|c|c|c}
        \hline
        RJA & $L_{ctr}$ & $L^{crf}_{ce}$ & Yelp (1\%) & T-Soc (0.01\%) & DGraph (1\%)\\
        \hline
        \xmark & \xmark & \xmark & $72.88_{\pm0.82}$ & $81.58_{\pm1.95}$ & $73.09_{\pm0.57}$\\
        \cdashline{1-6}
        \xmark & \xmark & Mix-up & $72.17_{\pm1.08}$ & $83.15_{\pm1.31}$ & $72.25_{\pm0.76}$\\
        \xmark & \xmark & \cmark & $73.57_{\pm1.01}$ & $87.48_{\pm1.77}$ & $73.36_{\pm0.63}$\\
        \xmark & \cmark & \xmark & $74.05_{\pm0.93}$ & $83.96_{\pm2.00}$ & $72.90_{\pm0.51}$\\
        \xmark & \cmark & \cmark & $73.94_{\pm1.41}$ & $88.54_{\pm1.46}$ & $73.51_{\pm0.51}$\\
        \cdashline{1-6}
        \cmark & \xmark & \xmark & $75.89_{\pm1.22}$ & $76.91_{\pm3.25}$ & $76.44_{\pm0.63}$\\
        \cmark & \xmark & Mix-up & $75.83_{\pm1.44}$ & $79.36{\pm2.46}$ & $75.15_{\pm0.53}$\\
        \cmark & \xmark & \cmark & $78.69_{\pm1.29}$ & $85.01_{\pm2.45}$ & $76.54_{\pm0.59}$\\
        \cmark & \cmark & \xmark & $77.35_{\pm0.96}$ & $81.45_{\pm2.54}$ & $76.28_{\pm0.63}$\\
        \cdashline{1-6}
        \cmark & \cmark & \cmark & $79.60_{\pm0.74}$ & $88.79_{\pm1.35}$ & $76.62_{\pm0.54}$\\
        \hline
    \end{tabular}
    }
    \label{tab:supp_exp:abs_sage}
\end{table}

We show the completed experimental results of the ablation studies on CRoC(GIN) and CRoC(SAGE) in Table \ref{tab:supp_exp:abs_gin} and Table \ref{tab:supp_exp:abs_sage}, respectively.
We additionally replace context-refactoring with Mix-up \cite{mixup} for comparison.
From the results, we can find that introducing Mix-up can hardly improve the model's performance.
This is because the input space of the Mix-up variant is essentially the convex hull of the labeled training set, while the size of labeled training set is extremely small in these three datasets. 
Applying Mix-up in a scenario with such limited supervision may attenuate the signals from the target of interest (i.e., anomalous nodes), making it hard to distinguish informative patterns for GAD.
We can also observe that the Relation-aware Joint Aggregation (RJA) is effective for all three large-scale datasets.
However, without context refactoring or the contrastive learning scheme, which enables the model to learn from the abundant unlabeled data, individually introducing RJA into the model may lead to overfitting when available training labels are very few (e.g., in T-Soc) in the graph.
On the other hand, when all modules are implemented, CRoC(SAGE) boosts the backbone model by a large margin.
This fact demonstrates that all proposed modules can complement each other and synergize well for GAD tasks.

\subsection{Qualitative Analyses and Visualizations}
\label{app:supp_result:qua_anal}

\begin{figure}[tbp]
  \centering
  \includegraphics[width=0.5\textwidth]{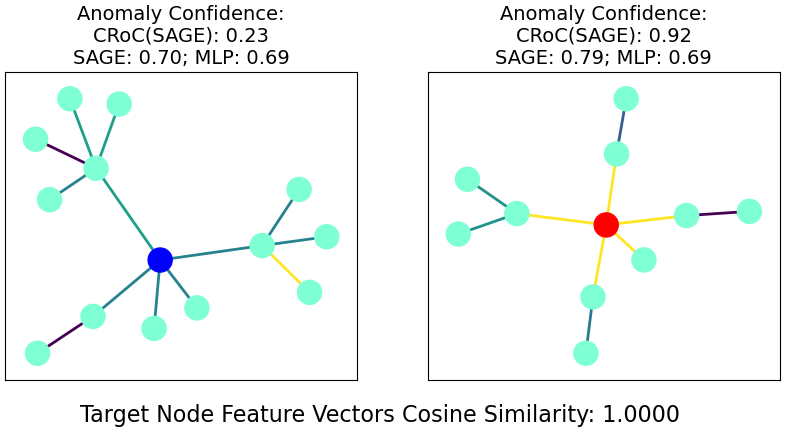}
  \includegraphics[width=0.5\textwidth]{supp_figs/RJA_cases/sim_1273_node_597644_1.00.png}
  \includegraphics[width=0.5\textwidth]{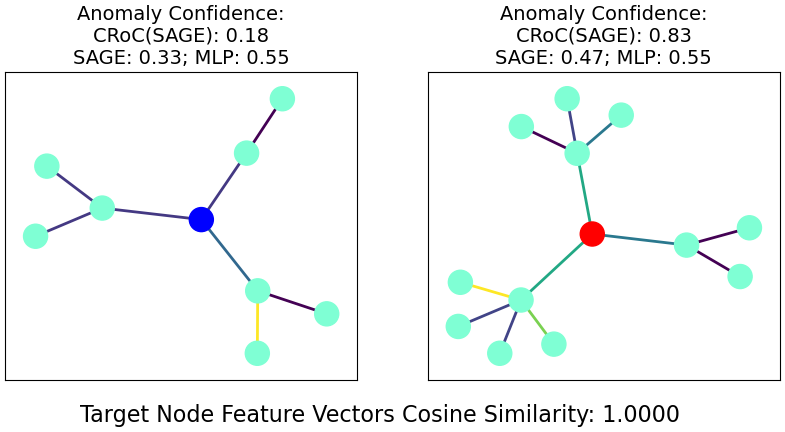}
  \includegraphics[width=0.5\textwidth]{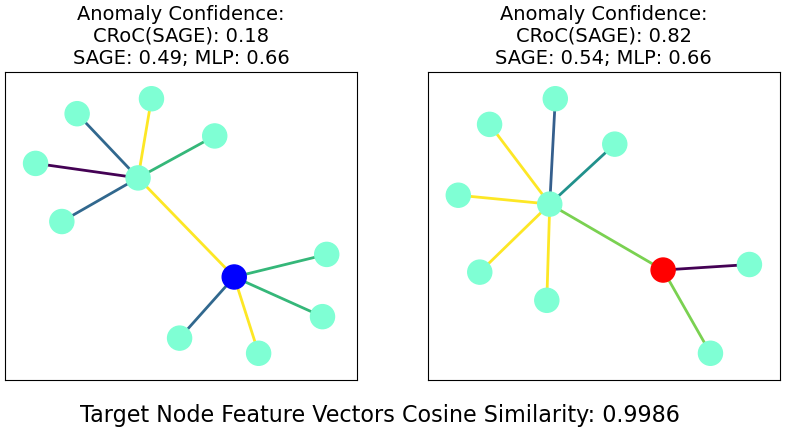}
  
  \caption{Visualization of the local graph topology of some typical cases in DGraphFin. 
  Blue nodes and red nodes indicate the target benign and anomalous instances, respectively. 
  Green nodes are neighbors in the induced computational graphs of the corresponding target node. 
  Different edges' colors indicate different types of interaction relations. 
  The value of anomaly confidence corresponds to Equ (\ref{equ:method:pred}) in our main text.}
  \label{fig:supp:rja_viz}
\end{figure}

Quantitative experiment results in the main texts and supplementary ablation studies have shown the capability of CRoC in modeling multi-relation graphs for GAD tasks.
In this section, we visualize some cases in DGraphFin to further demonstrate its effectiveness in enhancing the representative power of the backbone GNN model.
In Fig. \ref{fig:supp:rja_viz}, we show the local graph topology of four node pairs for comparison.
The target nodes are either marked in blue (benign users) or red (anomalous users).
The green nodes are neighbors in the local graph.
Different edge colors correspond to different interaction relations.
We also list the cosine similarity of the raw features of each pair of target nodes and their predictions yielded by MLP, SAGE, and CRoC(SAGE).

We can observe that each pair of target nodes has similar raw node features (cosine similarity is close to $1$).
The local graph structure is also similar within each pair.
This is common for camouflaged instances, where illicit users may pretend to be normal by declaring a personal profile similar to benign users (feature camouflage) or act (interact) with others like normal users (behavior camouflage).
Therefore, some detectors relying solely on node features (such as MLP or XGBoost) will fail to distinguish feature camouflage cases from those of normal users.
GNN models, which benefit from learning different graph structural patterns, can perform better against feature camouflage.
However, typical GNNs (such as GCN and GraphSAGE) are relation-aware.
When the local graph structure of the anomalous node is similar to that of a normal node, these models may find it challenging to distinguish behavior camouflage cases.

In contrast, CRoC(SAGE) variants can differentiate these cases well.
Note that CRoC(SAGE) is trained not only on the original graph but also on the context-refactored graph, which deliberately simulates node feature camouflage in the GAD context.
This induces the backbone model to be adaptive to feature camouflage cases.
In addition, as shown in Fig. \ref{fig:supp:rja_viz}, even if the local graph structure of a malicious node is similar to that of a benign node, the relations included in the two graphs should not be identical.
In CRoC, Relation-aware Joint Aggregation (RJA) endows the backbone GNN model with the power to differentiate various relations during the aggregation process.
In other words, RJA enables the model to mine more discriminative clues provided by edges, which drives a GNN to learn more powerful representations for nodes.
This makes it possible for the GNN model to distinguish behavior camouflage cases from normal cases, which effectively improves the performance of the GAD task.

\subsection{Performance under Different Training Label Rate Settings}
\label{app:supp_result:label_rate}

In this part, we investigate the performance of the proposed method under different training label rate settings.
In the experiments, we use the GIN model as the backbone and evaluate it on Yelp and T-Soc.
Specifically, we randomly sample $\{ 40\%, 20\%\, 10\%, 1\%, 0.1\% \}$ of nodes from the graph to form the labeled training set.
The remaining nodes are split into validation and test sets, with a ratio of $1:2$.
We also evaluate two state-of-the-art GAD models, i.e., BWGNN \cite{bwgnn} and GHRN \cite{ghrn}, for comparison.
All other settings are the same as those in \ref{app:exp_setting}.

\begin{figure}[tp]
    \centering
    \subfigure[Results on Yelp.]{\includegraphics[width=0.5\textwidth]{supp_figs/vary_splits/diff_label_rate_Yelp.png}}
    \subfigure[Results on T-Soc.]{\includegraphics[width=0.5\textwidth]{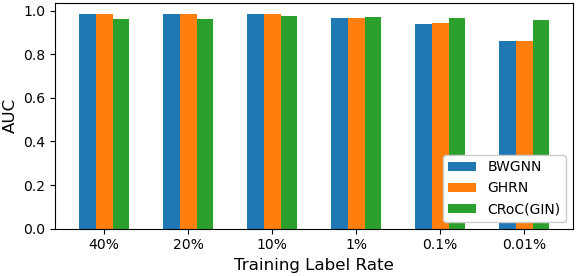}}
        
    \caption{
        Experimental results for different training label rate settings.
        }
    \label{fig:supp_exp:diff_tr_rate}
\end{figure}

Experimental results are shown in Fig. \ref{fig:supp_exp:diff_tr_rate}.
We can observe that CRoC(GIN) still outperforms BWGNN and GHRN in Yelp and is comparable to them in T-Soc when labels are sufficient for training (i.e., $40\%$, $20\%$, and $10\%$ training label rate settings).
However, when we reduce the number of labels available for training, the AUC metric of BWGNN and GHRN drops dramatically.
In contrast, CRoC(GIN) still performs relatively well under the same limited supervision scenarios.
The performance gap between CRoC(GIN) and the other two methods becomes more significant when the training label set is smaller.
Such observations demonstrate that CRoC is effective for scenarios with different levels of supervision, and it shows great superiority over other methods when a smaller annotation budget (e.g., the number of available training labels) is provided.

\subsection{Investigations on Hyper-parameters}
\label{app:supp_result:hp_inves}

In this part, we investigate the sensitivity of the tunable hyper-parameters, including $\alpha$ (refer to Equ (\ref{equ:method:feat_shuf}) in the main text), $\gamma$, and $\eta$ (refer to Equ (\ref{equ:method:overall_loss}) in the main text).

\subsubsection{The Impact of Native Node Features}

In Equ (\ref{equ:method:feat_mix}) in the main text, we use $\alpha$ to control the importance of native node features for context refactoring.
In this section, we investigate the impact of $\alpha$ for the enhancement effect.
Specifically, we use SAGE \cite{graphsage} as the backbone and vary $\alpha$ in $\{ 0, 0.25, 0.5, 0.75, 0.9 \}$.
The experiments are conducted on a small-scale dataset (T-Fin), a medium-scale dataset (Yelp), and a large-scale dataset (DGraph).
All other hyper-parameters follow the best configurations shown in Table \ref{tab:supp_exp:hp_details}.

\begin{figure}
    \centering
    \includegraphics[width=\linewidth]{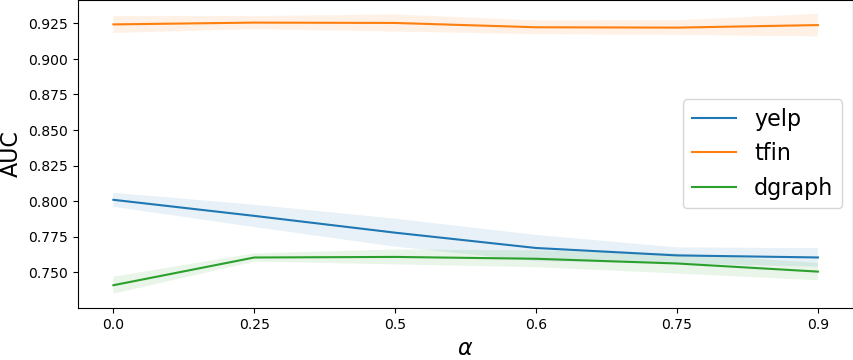}
    \caption{The results of varying the value of $\alpha$ for CRoC(SAGE).}
    \label{fig:supp_exp:vary_alpha}
\end{figure}

From Fig. \ref{fig:supp_exp:vary_alpha}, we can observe that CRoC(GIN) performs stably in T-Fin as the value of $\alpha$ varies, while the AUC shows a different trend in Yelp and DGraph.
This may indicate that there are fewer camouflaged cases in T-Fin, or there are some critical features that can distinguish the anomalous instance, since feature-based common classifiers (such as XGBoost) can already achieve good performance in T-Fin (refer to Table 1 in our main text).
In contrast, camouflaged cases may be more common to be found in Yelp and DGraph, where the value of $\alpha$ makes a difference for the prediction.
In other words, the importance of $\alpha$ to the overall performance may reflect the camouflage level of a dataset, which is promising to be left as a direction for future research.

\subsubsection{The Impact of Tunable Factors in the Loss Function}

In this part, we assess the impact of $\gamma$ and $\eta$ on the performance.
Basic configurations and settings are the same as above.

\begin{figure}
    \centering
    \includegraphics[width=1\linewidth]{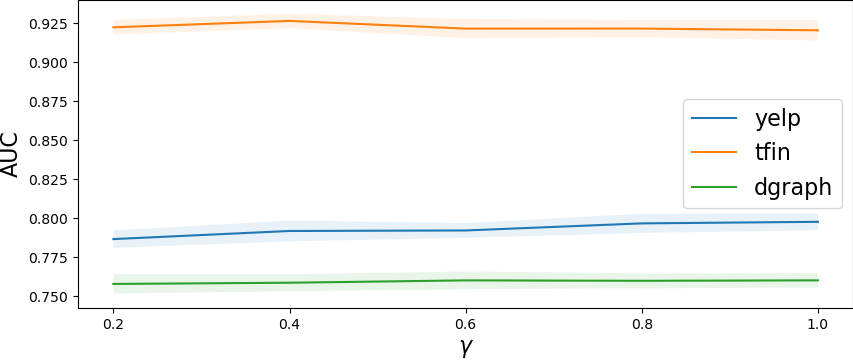}
    \caption{The results of varying $\gamma$ for CRoC(SAGE).}
    \label{fig:supp_exp:vary_gamma}
\end{figure}

\begin{figure}
    \centering
    \includegraphics[width=1\linewidth]{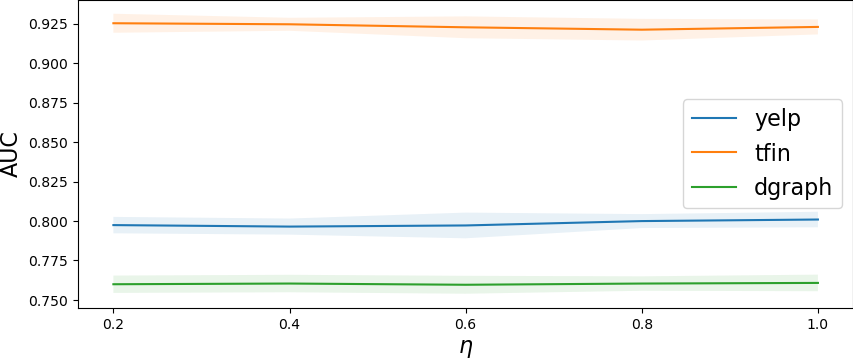}
    \caption{The results of varying $\eta$ for CRoC(SAGE).}
    \label{fig:supp_exp:vary_eta}
\end{figure}

Fig. \ref{fig:supp_exp:vary_gamma} and Fig. \ref{fig:supp_exp:vary_eta} shows the experimental results.
It can be observed that the AUC metric of CRoC(SAGE) doesn't fluctuate dramatically as the factor value varies.
This means that CRoC is robust for training, which benefits it from generating more consistent predictions for the GAD task.

\subsection{Comparisons with Specialized GAD Methods in Different Evaluation Protocols}
\label{app:supp_result:diff_proto}

In this part, we compare our proposed methods with more specialized GAD models.
However, due to the differences in evaluation protocols (e.g., different evaluation datasets, different metrics, different dataset split settings, etc.), it is hard to directly compare the performance of CRoC's variants with their publicly reported results.
Concretely, some specialized GAD models and their experimental settings are grouped as follows.

\begin{enumerate}
    \item ADA-GAD \cite{ada-gad}, GAD-NR \cite{gad-nr}, TAM \cite{tam} and GADAM \cite{GADAM_iclr}: this group of models learn to detect anomalous nodes in an unsupervised manner.
    The performance is measured by the AUC score.
    \item ARC \cite{arc}: this work is trained on several large-scale datasets and then tested on a set of smaller-scale datasets, including Weibo, Reddit, and Amazon.
    The performance is measured in the AUC score.
    \item GENGA and CGENGA \cite{genga}: these two methods (variants) are trained with a ratio of $20\%$ labeled nodes (i.e., $20\%$ of the nodes in a graph are labeled for training).
    The performance is measured by the AUC score.
    \item gADAM \cite{gADAM_cikm}: this work samples 50 labeled anomalies (labels) for training.
    The performance is measured by the AUC score.
    \item BSL \cite{bsl}: this work uses the same dataset split as our evaluation settings (i.e., $1\%$ label rate for T-Finance, Yelp, and Amazon, while $0.01\%$ for T-Social).
    However, it is measured by the Macro-F1 score.
\end{enumerate}

To compare with these methods, we conduct a series of experiments by aligning our settings with the aforementioned works.
Specifically, we evaluate CRoC(GIN) or CRoC(SAGE) on the same set of datasets, by using the same train/validation/test dataset splits and evaluation metrics as the models in each group.
Among the listed models, we replicate the results of ADA-GAD \cite{ada-gad}and GAD-NR \cite{gad-nr} using their public code repositories \footnote{https://github.com/jweihe/ADA-GAD} \footnote{https://github.com/Graph-COM/GAD-NR}.
For other methods, we compare CRoC (implemented in GIN) with the results reported in their published papers.
Note that for group 1 (unsupervised models), there are no explicit train/validation/test splits.
Therefore, we compare them with CRoC trained under the default split settings (i.e., $1\%$ of nodes (with labels) are sampled for training, and the remaining are split in a ratio of $1:2$ for validation and testing).
All other settings are kept the same as stated in \ref{app:exp_setting:config}.

\begin{table}[t]
    \centering
    \caption{Experimental results (in AUC) on five GAD datasets. CRoC(GIN) is trained under the $1\%$ label rate setting. (OOM: out of memory. N/A: not available in the published paper.)}
    \begin{tabular}{c|ccccc}
        \hline
        Dataset & T-Fin & Yelp & Amazon & Weibo & Reddit \\
        \hline
        ADA-GAD & OOM & OOM & OOM & $\bm{92.56}$ & $56.89$ \\
        GAD-NR & OOM & $54.36$ & $70.00$ & $87.71$ & $57.99$ \\
        TAM & $61.75$ & $56.43$ & $70.64$ & N/A & $60.23$ \\
        GADAM & N/A & N/A & N/A & N/A & 58.09 \\
        ARC & N/A & N/A & $80.67$ & $88.85$ & $\bm{60.04}$ \\
        \hdashline
        CRoC(GIN) & $\bm{91.21}$ & $\bm{81.64}$ & $\bm{89.65}$ & $91.52$ & $59.92$ \\
        \hline
    \end{tabular}
    \label{tab:supp_exp:comp_unsup}
\end{table}

\begin{table}[t]
    \centering
    \caption{Experimental results (in AUC) on three GAD datasets. CRoC(GIN) is trained under the $20\%$ label rate setting.}
    \begin{tabular}{c|ccc}
        \hline
        Dataset & Yelp & Weibo & Reddit \\
        \hline
        GENGA & $83.74$ & $90.02$ & $70.79$  \\
        CGENGA & $83.73$ & $89.44$ & $\bm{71.07}$ \\
        \hdashline
        CRoC(GIN) & $\bm{85.88}$ & $\bm{95.45}$ & $66.82$ \\
        \hline
    \end{tabular}
    \label{tab:supp_exp:comp_20lb}
\end{table}

\begin{table}[t]
    \centering
    \caption{Experimental results (in AUC) on Yelp and Amazon. CRoC(GIN) is trained using 50 anomalous labels and 50 normal labels.}
    \begin{tabular}{c|cc}
        \hline
        Dataset & Yelp & Amazon \\
        \hline
        gADAM & $71.2$ & $72.3$ \\
        \hdashline
        CRoC(GIN) & $\bm{76.17}$ & $\bm{82.16}$ \\
        \hline
    \end{tabular}
    \label{tab:supp_exp:comp_50-50}
\end{table}

\begin{table}[t]
    \centering
    \caption{Experimental results (\textcolor{red}{in Macro-F1}) on four GAD datasets. Label rate is marked in parentheses.}
    \resizebox{\columnwidth}{!}{
    \begin{tabular}{c|cccc}
        \hline
        Dataset &T-Soc ($0.01\%$) & T-Fin ($1\%$) & Yelp ($1\%$) & Amazon ($1\%$) \\
        \hline
        BSL & $75.67$ & $\bm{86.10}$ & $62.83$ & $\bm{91.42}$ \\
        \hdashline
        CRoC(GIN) & $\bm{80.80}$ & $84.26$ & $\bm{70.87}$ & $74.81$  \\
        \hline
    \end{tabular}
    }
    \label{tab:supp_exp:comp_macrof1}
\end{table}

The experimental results are shown from Table \ref{tab:supp_exp:comp_unsup} to \ref{tab:supp_exp:comp_macrof1}.
We can observe that CRoC(GIN) outperforms most of the cutting-edge GAD methods in most cases.
When the absolute number of labeled training samples is very small (e.g., in Table \ref{tab:supp_exp:comp_50-50}) or the training label rate setting is extremely low (e.g., in the T-Soc dataset in Table \ref{tab:supp_exp:comp_macrof1}), CRoC(GIN) shows much better performance, leading by a large margin.
In addition, CRoC(GIN) tends to work better in large-scale graphs, such as in Yelp (with over 45 thousand nodes) and T-Soc (with 5 million nodes), where it dominates all competitive methods across all different settings.
This may be owed to the context refactoring strategy and the introduction of the contrastive learning scheme, which actively incorporates unlabeled data into the training process and thereby helps learn more diverse knowledge from the graph.
All these results demonstrate that CRoC is adaptive to different kinds of scenarios, with decent generalization ability.






\bibliography{croc}

\end{document}